\documentclass[letterpaper, 10 pt, journal, twoside]{IEEEtran}

\usepackage{decar-common}
\usepackage{decar-dynamics}
\usepackage{decar-lie}
\usepackage{decar-post}
\usepackage{tabularx}
\usepackage{booktabs}

\usepackage{xfrac}
\newlength{\subfigheight}
\newsavebox{\subfigbox}

\Crefformat{figure}{#2Fig.~#1#3}

\makeatletter
\AtBeginDocument{
  \check@mathfonts
}
\makeatother

\DeclareMathOperator{\Exp}{Exp}

\newcolumntype{C}[1]{>{\centering\let\newline\\\arraybackslash\hspace{0pt}}m{#1}}
\newcolumntype{Y}{>{\centering\arraybackslash}X}
\newcolumntype{M}[1]{>{\centering\arraybackslash}m{#1}}

\begin{document}

% ensure all superscripts / subscripts at the same elevation
\fontdimen16\textfont2=\fontdimen17\textfont2
\fontdimen13\textfont2=5pt

% title
\title{An $SE(3)$ Noise Model for Range-Azimuth-Elevation Sensors}

% author
\author{Thomas~Hitchcox~and~James~Richard~Forbes,~\IEEEmembership{Member,~IEEE}%
        
% thanks block
% TODO Include manuscript received, etc.
\thanks{This work was supported by Voyis Imaging Inc. through the Natural
Sciences and Engineering Research Council of Canada (NSERC) Collaborative
Research and Development (CRD) program, and the McGill Engineering Doctoral
Award (MEDA) program.}%

% author attribution
\thanks{Thomas~Hitchcox and James~Richard~Forbes are with the Department of Mechanical
Engineering, McGill University, Montreal, Quebec H3A~0C3, Canada.
\texttt{thomas.hitchcox@mail.mcgill.ca, james.richard.forbes@mcgill.ca}.}%

% received, revised, etc.
% TODO uncomment
% \thanks{Digital Object Identifier (DOI): see top of this page.}
}%

% page header
% TODO update header
% \markboth{IEEE Robotics and Automation Letters. Preprint version. Accepted
% XXX, 2024}{Hitchcox and Forbes: An $SE(3)$ Noise Model for Range-Azimuth-Elevation Sensors}%
\markboth{}{Hitchcox
and Forbes: An $SE(3)$ Noise Model for Range-Azimuth-Elevation Sensors}%

% make the title area
\maketitle

% abstract
\begin{abstract}
    Scan matching is a widely used technique in state estimation.  Point-cloud alignment, one of the most popular methods for scan matching, is a weighted least-squares problem in which the weights are determined from the inverse covariance of the measured points.  An inaccurate representation of the covariance will affect the weighting of the least-squares problem.  For example, if ellipsoidal covariance bounds are used to approximate the curved, ``banana-shaped'' noise characteristics of many scanning sensors, the weighting in the least-squares problem may be overconfident.  Additionally, sensor-to-vehicle extrinsic uncertainty and odometry uncertainty during submap formation are two sources of uncertainty that are often overlooked in scan matching applications, also likely contributing to overconfidence on the scan matching estimate.  This paper attempts to address these issues by developing a model for range-azimuth-elevation sensors on matrix Lie groups.  The model allows for the seamless incorporation of extrinsic and odometry uncertainty.  Illustrative results are shown both for a simulated example and for a real point-cloud submap collected with an underwater laser scanner.
\end{abstract}

% keywords

% -----------------------------------------------------------
% -----------------------------------------------------------
\section{Introduction}
\label{sec:intro}

\IEEEPARstart{S}{} can matching, where a relative pose is computed by matching and aligning scan data, is a widely used technique in state estimation applications.  Scan data is often collected using radar
\cite{Retan2022,Lisus2023,Xu2024}, lidar \cite{Zhang2014,Ye2019a}, or laser
scanners \cite{Palomer2018,Palomer2019,Hitchcox2023a}.  Each of these sensors
may be represented using a range-azimuth-elevation (RAE) sensor model
\cite[\textsection6.4.3]{Barfoot2017a}.

A widely-used method of scan matching is point-cloud alignment, in which a relative pose estimate is obtained by aligning two sets of 3D point
measurements.  Point-cloud alignment algorithms such as Horn's method
\cite{Horn1987,Arun1987} and the WOLATE algorithm \cite{Qian2018} minimize a sum
of weighted squared errors between associated points, in which the weight is generally the inverse of the point measurement covariances.  Many point-cloud
alignment algorithms model the point measurements as vectors in Euclidean space,
with ellipsoidal uncertainties.  However, RAE sensors fundamentally exhibit
curved, ``banana-shaped'' uncertainty envelopes \cite{Cesic2016,Retan2022}.
\textit{Linearization of the RAE sensor model generally leads to a loss of accuracy, as
ellipsoidal covariance bounds must then be used to represent fundamentally
nonlinear noise characteristics} \cite{Chirikjian2014}.  This can lead to an
overestimation of the point uncertainty at close ranges, and an underestimation
of point uncertainty at long ranges \cite{Retan2022}, which in turn leads to inaccurate weighting in the point-cloud alignment problem.  Additionally, point-cloud alignment is often preceded by a coarse alignment step to determine the point associations.  Inaccurate covariance bounds may also impact the coarse alignment step by leading to poor data association decisions \cite{Neira2001}.

Point-cloud alignment is known to be overconfident, meaning the error in the mean relative pose estimate is inconsistent with the accompanying uncertainty bounds \cite{Bonnabel2016}.  Two known sources of this inconsistency are the effects of initialization (effectively, data association \cite{Olson2009}) and bias in the scanning sensor \cite{Brossard2020}.  However, two additional sources of uncertainty that are often overlooked in scan matching applications are uncertainty in the sensor-to-vehicle extrinsic alignment and uncertainty in the vehicle odometry.  Odometry uncertainty is relevant in the case where multiple scan measurements are combined into ``submaps'' in order to facilitate scan matching.  For example, this is commonly done with laser scanners and multibeam sonar \cite{Barkby2012}.  The relative position of scanned points at opposite ends of a submap will be more uncertain than the relative position of nearby points leading to a heterogeneous noise distribution across the domain of the submap.  This will in turn influence point measurement weighting and ultimately the uncertainty of the scan matching estimate.

% -----------------------------------------------------------
% -----------------------------------------------------------
\subsection{Related Work}
\label{sec:relatedwork}

The standard RAE measurement model is prevalent in robotics, and few papers have
investigated alternative noise models for these sensors.  The ability of matrix
Lie group-based measurement models to represent fundamentally nonlinear
sensors, such as range-bearing sensors, was first acknowledged in
\cite{Chirikjian2014}.  The developed framework showed how to perform a
measurement update in a matrix Lie group-based estimator, without the need to
linearize the measurement model. Later, \cite{Cesic2016} modeled radar and
stereo camera measurements on the matrix Lie group ${SO(2) \times \rnums}$,
demonstrating how the measurement model naturally captures the ``banana-shaped''
noise distribution of these sensors. Attitude measurements were parameterized
directly on $SO(3)$ in \cite{Lee2018}, which was later augmented to ${SO(3)
\times \rnums^n}$ in order to incorporate gyro biases \cite{Wang2021}.  Recently,
\cite{Xu2024} presented an ${\rnums \times \mathbb{S}^2}$ model for radar
sensors, which models azimuth and elevation uncertainty on the tangent plane of
the sphere $\mathbb{S}^2$.  This model is capable of representing curved
uncertainty envelopes for RAE measurements, which are then used in a Mahalanobis
distance test for both matching likely pairs of landmarks and rejecting outlier
matches between consecutive radar scans.

% -----------------------------------------------------------
% -----------------------------------------------------------
\subsection{Contribution}
\label{sec:contribution}

The main contribution of this paper is the development of a nonlinear noise
model for RAE sensors within a matrix Lie group framework.  The use of this
framework allows for the seamless incorporation of uncertainty in the
sensor-to-vehicle extrinsic estimate, and even for the incorporation of
uncertainty accumulated in the vehicle pose estimate during scan collection.
This can be especially important for situations where 3D ``submaps'' are built
up from multiple 2D scan profiles, for example when using multibeam sonar
\cite{Barkby2012} or ``push-broom'' laser scanners \cite{Hitchcox2023a}.  

The rest of this paper is organized as follows.  \Cref{sec:prelim} introduces the notation and matrix Lie group concepts used throughout the paper.  \Cref{sec:methodology} develops the methodology, including the RAE sensor model and the incorporation of extrinsic and odometry uncertainty into the submap noise profile.  Simulated and field results are presented in \Cref{sec:results}.  The paper concludes in \Cref{sec:conclusion} with a short summary and proposals for future work.

% -----------------------------------------------------------
% -----------------------------------------------------------
\section{Preliminaries}
\label{sec:prelim}

This section reviews geometry and uncertainty on matrix Lie groups, and
introduces the notation used in this paper. 

% -----------------------------------------------------------
% -----------------------------------------------------------
\subsection{Notation}
\label{sec:frames}

\begin{figure}
	\centering
	\includegraphics[width=0.95\columnwidth]{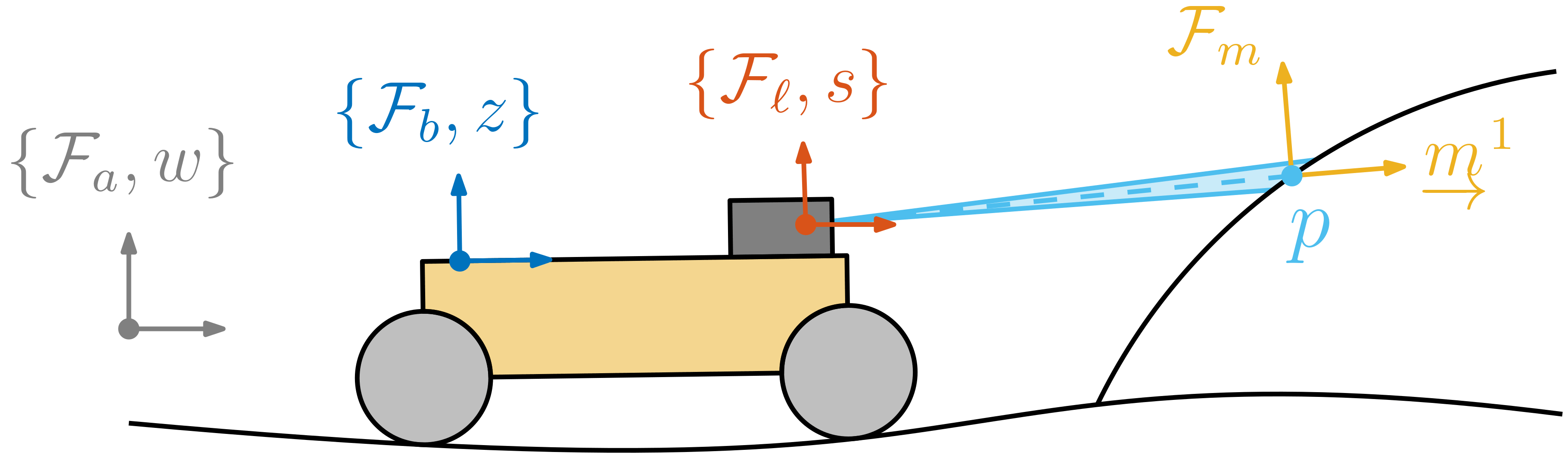}
	\caption{Reference frames and datums for the problem.  From left to right,
	the labels indicate the \colour{matlab_grey}{world frame}, the
	\colour{matlab_blue}{vehicle frame}, and the \colour{matlab_orange}{sensor
	frame} (and their respective datums), as well as the
	\colour{matlab_cyan}{measured point $p$}.  The first basis vector $\ura{m}^1$ of the
	\colour{matlab_yellow}{measurement-aligned frame} is also indicated.}
	\label{fig:frames_and_datums}
\end{figure}

A 3D dextral reference frame $\rframe{a}$ is composed of three
orthonormal physical basis vectors $\{ \ura{a}^i \}^{3}_{i=1}$.  The position of point $z$ relative to point $w$ is resolved in $\rframe{a}$ as ${\mbf{r}^{zw}_a \in \rnums^3}$ and in frame
$\rframe{b}$ as $\mbf{r}^{zw}_b$.  These quantities are related via
${\mbf{r}^{zw}_a = \mbf{C}_{ab} \mbf{r}^{zw}_b}$, with $\mbf{C}_{ab}$ a
direction cosine matrix (DCM), ${\mbf{C} \in SO(3) = \{ \mbf{C} \in
\rnums^{3\times3} \, | \, \mbf{C}\mbf{C}^\trans = \eye, \det \mbf{C} = +1 \}}$
\cite[\textsection7.1.1]{Barfoot2017a}.  Time-varying quantities are indicated
by the subscript $(\cdot)_k$, for example $\mbf{r}^{z_kw}_a$ describes the
position of moving point $z$ at time $t_k$.  In this work, point $w$ is the
world datum, point $z$ is the vehicle datum, point $s$ is the sensor datum, and
point $p$ is the scanned point.  Furthermore, $\rframe{a}$ is the local geodetic
reference frame, $\rframe{b}$ is the vehicle frame, $\rframe{\ell}$ is the
sensor frame (taken to be a laser scanner or lidar), and $\rframe{m}$ is the ``measurement aligned'' frame used for
defining the measurement covariance.  A summary of the various datums and reference
frames is shown in \Cref{fig:frames_and_datums}.

% -----------------------------------------------------------
% -----------------------------------------------------------
\subsection{Uncertainty on Matrix Lie Groups}
\label{sec:liegroups}

A vehicle's position and orientation, referred to collectively as the vehicle's
\textit{pose}, is often parameterized as an element of a matrix Lie group $G$.
Pose uncertainty is modeled in the matrix Lie algebra, defined as the tangent
space at the group identity, ${\mathfrak{g} \triangleq T_\eye G}$
\cite{Sola2018}.  Given a perturbation ${\mbf{T} = \mbfbar{T} \exp (\delta
\mbs{\xi}^\wedge)}$, with ${\delta \mbs{\xi}^\wedge \in \mathfrak{g}}$ and
${\delta \mbs{\xi} \in \rnums^d}$, the uncertainty on matrix Lie group element
$\mbf{T}$ is described by ${\delta \mbs{\xi} \sim \mathcal{N} (\mbf{0},
\mbs{\Sigma})}$, ${\mbs{\Sigma} = \expect{ \delta \mbs{\xi} \, \delta
\mbs{\xi}^\trans}}$.  Note the operator ${\Exp(\cdot) : G \to \rnums^d,
\Exp(\cdot) \triangleq \exp((\cdot)^\vee)}$, with ${(\cdot)^\vee : \mathfrak{g}
\to \rnums^d}$, will be used going forward to simplify expressions involving the
exponential map.  Finally, the adjoint \textit{operator}
${\text{Ad}_\mbf{T}(\cdot) : \mathfrak{g} \to \mathfrak{g}}$ maps perturbations
applied in the matrix Lie algebra to a location $\mbf{T}$ on the group, 
\begin{equation}
    \text{Ad}_\mbf{T}(\delta \mbs{\xi}^\wedge) \triangleq \mbf{T} \delta \mbs{\xi}^\wedge \mbf{T}\inv.
\end{equation}
The adjoint \textit{matrix} encodes the effects of the adjoint operator directly
in $\rnums^d$, 
\begin{equation}
    \Adj(\mbf{T}) \delta \mbs{\xi} \triangleq \left( \mbf{T} \delta \mbs{\xi}^\wedge \mbf{T} \right)^\vee.
\end{equation}
The adjoint matrix appears in expressions of uncertainty over compound, inverse,
and relative poses \cite{Mangelson2019}.  

This work uses both matrix Lie group $SE(2)$
\cite[\textsection10.6.2]{Chirikjian2011} to represent 2D planar pose,
\begin{subequations}
    \begin{align}
        \mbf{T}^{zw}_{ab}(\mbf{C}_{ab}, \mbf{r}^{zw}_a) &= \begin{bmatrix}
            \mbf{C}_{ab}(\theta_{ba}) & \mbf{r}^{zw}_a \\ \mbf{0} & 1
        \end{bmatrix} \in SE(2), 
        \label{eqn:elementSE2} \\
        SE(2) &= \left\lbrace \mbf{T} \in \rnums^{3 \times 3} \, \big| \, \mbf{C} \in SO(2), \mbf{r} \in \rnums^2 \right\rbrace,
    \end{align}
\end{subequations}
and $SE(3)$ \cite[\textsection7.1.1]{Barfoot2017a} to represent general 3D pose,
\begin{subequations}
    \begin{align}
        \mbf{T}^{zw}_{ab}(\mbf{C}_{ab}, \mbf{r}^{zw}_a) &= \begin{bmatrix}
            \mbf{C}_{ab}(\mbs{\phi}_{ba}) & \mbf{r}^{zw}_a \\ \mbf{0} & 1
        \end{bmatrix} \in SE(3),
        \label{eqn:elementSE3} \\
        SE(3) &= \left\lbrace \mbf{T} \in \rnums^{4 \times 4} \, \big| \, \mbf{C} \in SO(3), \mbf{r} \in \rnums^3 \right\rbrace. 
    \end{align}
\end{subequations}
For $SE(2)$, the pose perturbation ${\delta \mbs{\xi}_{SE(2)} \in \rnums^3}$ and
adjoint matrix are given by, respectively, 
\begin{equation}
    \delta \mbs{\xi}_{SE(2)}^\wedge = \begin{bmatrix}
        \theta \\ \mbs{\rho}
    \end{bmatrix}^\wedge \in \mathfrak{se}(2), \quad \Adj(\mbf{T})_{SE(2)} = \begin{bmatrix}
        1 & \mbf{0} \\ -\mbs{\Omega} \mbf{r} & \mbf{C}
    \end{bmatrix},
\end{equation}
with
\begin{equation}
    \mbs{\Omega} = \begin{bmatrix}
        0 & -1 \\ 1 & 0
    \end{bmatrix}.
\end{equation}
For $SE(3)$, the pose perturbation ${\delta \mbs{\xi}_{SE(3)} \in \rnums^6}$ and
adjoint matrix are given by, respectively, 
\begin{equation}
    \delta \mbs{\xi}_{SE(3)}^\wedge = \begin{bmatrix}
        \mbs{\phi} \\ \mbs{\rho}
    \end{bmatrix}^\wedge \in \mathfrak{se}(3), \quad \Adj(\mbf{T})_{SE(3)} = \begin{bmatrix}
        \mbf{C} & \mbf{0} \\ \mbf{r}^\times \mbf{C} & \mbf{C}
    \end{bmatrix},
    \label{eqn:adjSE3}
\end{equation}
with $(\cdot)^\times$ the skew-symmetric operator
\cite[\textsection7.1.2]{Barfoot2017a}.

% -----------------------------------------------------------
% -----------------------------------------------------------
\section{Methodology}
\label{sec:methodology}

This section develops a noise model for RAE sensors on matrix Lie group $SE(3)$.
The section begins with a motivating example on $SE(2)$, in which a 2D point
cloud ``submap'' is formed from multiple range-bearing measurements.  An RAE
noise model for the aggregated submap is then developed to account for
measurement uncertainty, sensor-to-vehicle extrinsic uncertainty, and
uncertainty accumulated in the vehicle trajectory estimate during submap
formation.

% -----------------------------------------------------------
% -----------------------------------------------------------
\subsection{Motivating Example}
\label{sec:motivatingexample}

\begin{figure}
	\centering
	\includegraphics[width=0.95\columnwidth]{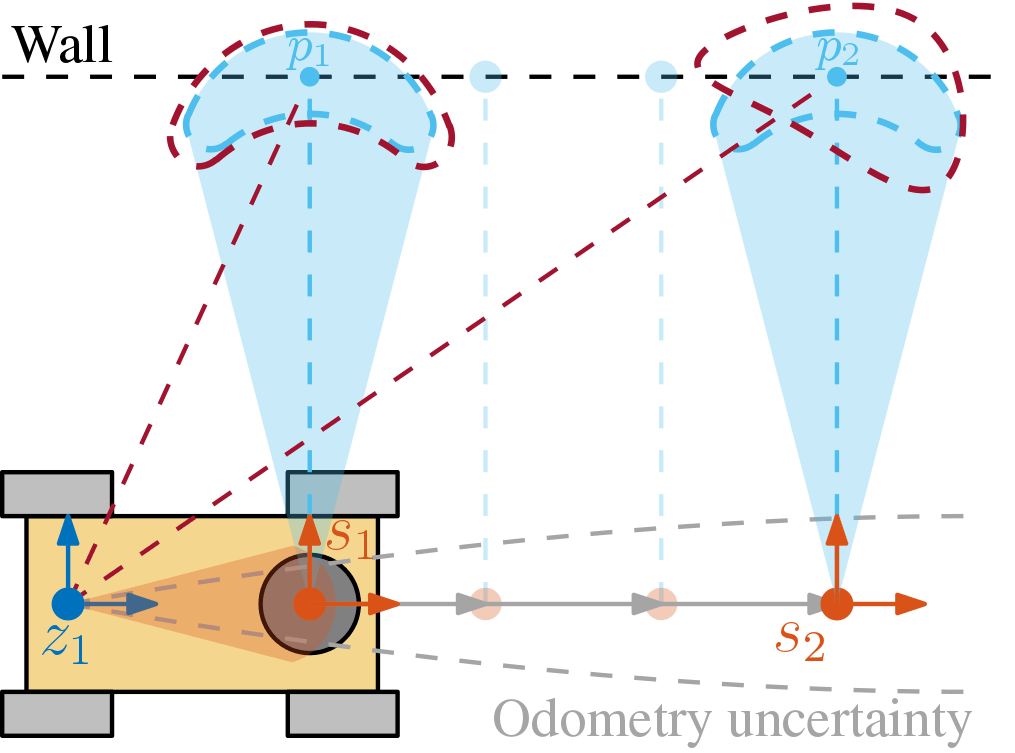}
	\caption{A simple motivating example on $SE(2)$, showing a top-down view of
	the vehicle from \Cref{fig:frames_and_datums} as it moves from left to
	right.  At each time step the vehicle records a single
	\colour{matlab_cyan}{range-bearing} measurement of a wall.  To perform scan
	matching, all measurements must be combined into a
	\colour{matlab_maroon}{submap} and therefore must be expressed relative to a
	single vehicle pose, arbitrarily chosen here to be
	\colour{matlab_blue}{$\mbf{T}^{z_1w}_{ab_1}$}.  The
	\colour{matlab_maroon}{noise profile of the aggregated submap} should
	reflect (a) \colour{matlab_cyan}{measurement uncertainty}, (b)
	\colour{matlab_orange}{sensor-to-vehicle extrinsic uncertainty}, and (c)
	\colour{matlab_grey}{odometry uncertainty} accumulated throughout submap
	formation.}
	\label{fig:motivating_example}
\end{figure}

Consider \Cref{fig:motivating_example}, in which the vehicle from
\Cref{fig:frames_and_datums} moves from left to right whilst recording
range-bearing measurements of a wall.  Individual measurements are then
aggregated into a more informative ``submap'' to facilitate scan matching.  

Scan matching algorithms, for example
\cite{Horn1987,Arun1987,Biber2003,Qian2018}, generally seek to find the rigid-body
transformation between two vehicle poses.  Therefore, each submap must first be
expressed relative to a single ``central'' vehicle pose.  Significant
uncertainty in the vehicle odometry estimate can accumulate during submap
formation (see \Cref{fig:motivating_example}), particularly if the submap is
large and the interoceptive sensors on the vehicle are of poor quality.
\textit{This uncertainty, together with the measurement uncertainty and the
uncertainty in the sensor-to-vehicle extrinsic estimate, should be reflected
in the submap noise model}.  Qualitatively, this means that point measurements
further from the selected ``central'' submap pose will have a larger covariance
than point measurements closer to the centre, and will therefore receive a lower
weight during scan matching optimization.  Through a more accurate accounting of
the submap measurement covariance, a comprehensive noise model based on matrix
Lie groups may help to address issues of overconfidence in scan matching
applications \cite{Censi2007,Brossard2020}.

% -----------------------------------------------------------
% -----------------------------------------------------------
\subsection{Part I: Defining a Noise Model for RAE Sensors}
\label{sec:sensormodel}

This section first derives a measurement model for RAE sensors, using the
machinery of matrix Lie groups to capture the ``banana-shaped'' noise
distribution known to be inherent to this class of sensor \cite{Chirikjian2014}.
For simplicity and ease of visualization, the derivation will begin on $SE(2)$,
that is, by first considering range-bearing (RB) measurements.

\begin{figure}
	\centering
	\includegraphics[width=0.95\columnwidth]{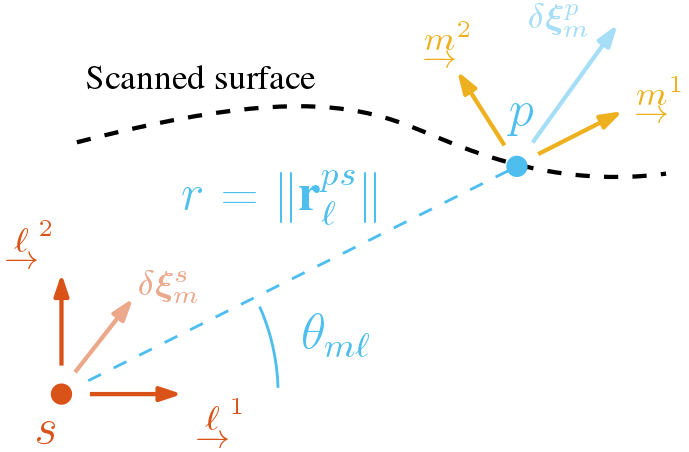}
	\caption{Parameterizing the range-bearing measurement ${\mbs{y} =
	\left\lbrace r, \theta_{m\ell} \right\rbrace }$ as an element of $SE(2)$.
	The adjoint matrix then provides a straightforward mapping between the
	sensor uncertainty \colour{matlab_orange_transparent}{$\mbs{\Sigma}^s_m$} and the measurement uncertainty
	\colour{matlab_cyan_transparent}{$\mbs{\Sigma}^p_m$}.  The
	\colour{matlab_yellow}{measurement-aligned frame} is also shown.}
	\label{fig:rangebearingSE2}
\end{figure}

\begin{figure}
	% preliminary
	\sbox\subfigbox{%
	  \resizebox{\dimexpr0.95\columnwidth-1em}{!}{%
		\includegraphics[height=3cm]{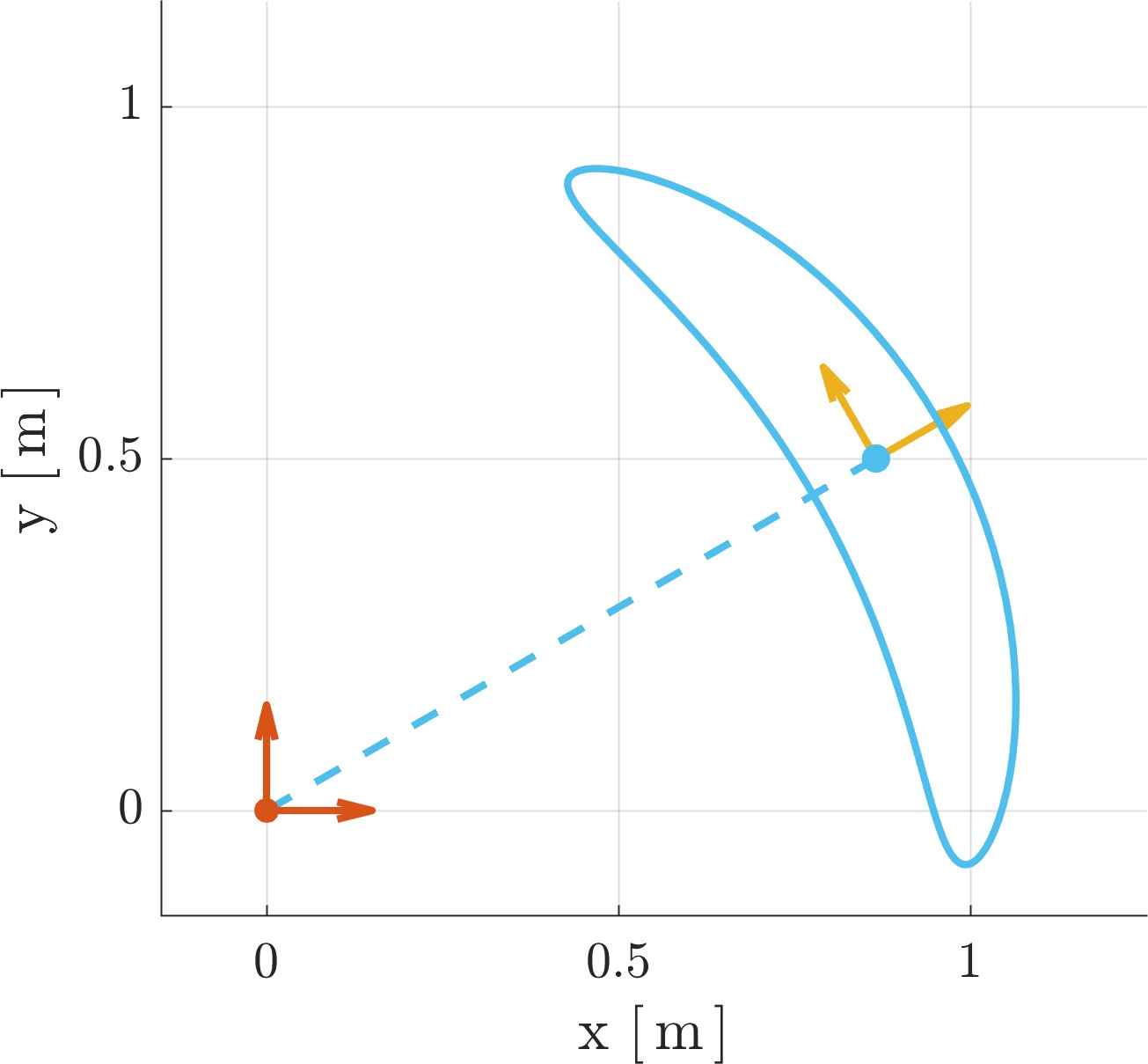}%
		\includegraphics[height=3cm]{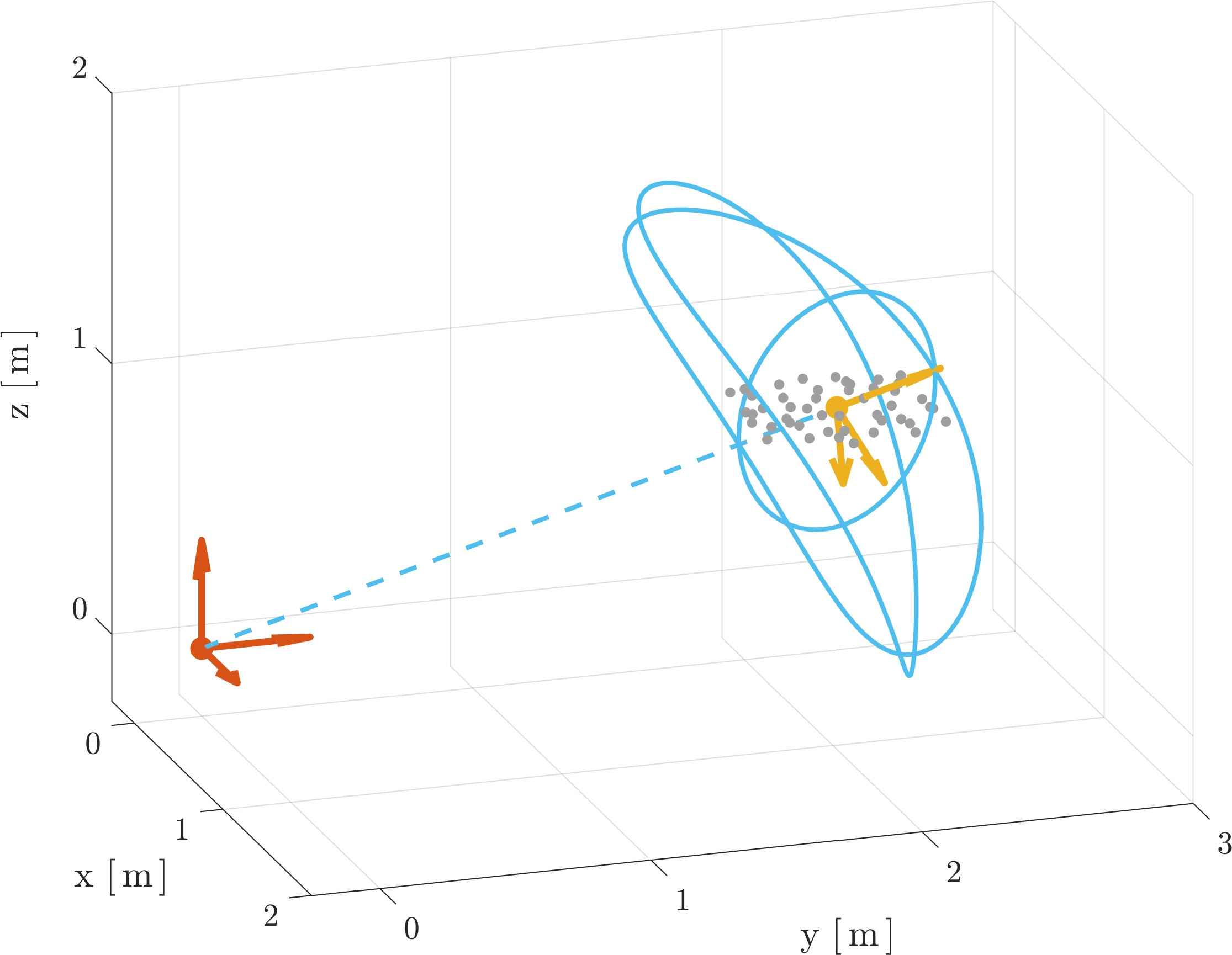}%
	  }%
	}
	\setlength{\subfigheight}{\ht\subfigbox}
	% the figure
	\centering
	\subcaptionbox{The RB measurement from \Cref{fig:rangebearingSE2}, with
	${\sigma_\textrm{r} = \SI{3}{\centi\meter}}$ and ${\sigma_\theta =
	\SI{10}{\deg}}$ (see \eqref{eqn:noise_cov_SE2}).
	\label{fig:range_bearing_matlab}}{%
	  \includegraphics[height=\subfigheight]{figs/range_bearing_SE2_matlab.jpg}
	}
	\hspace{3pt}
	\subcaptionbox{Simulated RAE measurement, ${\sigma_\textrm{r} =
	\SI{1}{\centi\meter}}$, ${\sigma_\phi = \SI{3}{\deg}}$, and ${\sigma_\theta
	= \SI{7}{\deg}}$ (see \eqref{eqn:noise_cov_SE3}). \label{fig:rae_matlab}}{%
	  \includegraphics[height=\subfigheight]{figs/rae_SE3_matlab_triad.jpg}
	}
    \caption{Illustrating the 99.73\% (3$\sigma$) confidence envelopes for
    simulated range-bearing measurements (left) and range-azimuth-elevation
    measurements (right).}
    \label{fig:matlab_illustrations}
\end{figure}

Consider \Cref{fig:rangebearingSE2}, where the range-bearing measurement of
point $p$ is defined by ${\mbs{y} = \left\lbrace r, \theta \right\rbrace}$, with
${r = \| \mbf{r}^{ps}_\ell \|}$.  Define the \textit{measurement-aligned frame}
$\rframe{m}$ such that $\ura{m}^1$ lies along the range direction, as shown in
yellow in \Cref{fig:rangebearingSE2}.  Now ${\theta = \theta_{m\ell}}$.  This
allows the range-bearing measurement to be expressed as an element of $SE(2)$,
\begin{equation}
    \mbf{T}^{ps}_{\ell m} = \begin{bmatrix}
        \mbf{C}_{\ell m}(\theta_{m\ell}) & \mbf{r}^{ps}_\ell(r) \\ \mbf{0} & 1
    \end{bmatrix} \in SE(2),
    \label{eqn:rangebearingSE2mean}
\end{equation}
in which ${\mbf{r}^{ps}_\ell(r) = \mbf{C}_{\ell m} \mbf{r}^{ps}_m}$,
${\mbf{r}^{ps}_m = \begin{matrix} [ \, r & 0 \, ] \end{matrix}^\trans}$.  From
\Cref{fig:rangebearingSE2}, the covariance on the RB \textit{sensor} may be
neatly expressed as 
\begin{equation}
	\mbs{\Sigma}^s_m = \expect{\, \delta \mbs{\xi}^s_m \left( \delta \mbs{\xi}^s_m \right)^\trans \,} = \begin{bmatrix}
		\sigma^2_\theta & 0 & 0 \\
		0 & \sigma^2_\textrm{r} & 0 \\
		0 & 0 & \delta^2
	\end{bmatrix},
    \label{eqn:noise_cov_SE2}
\end{equation}
where $\sigma_\theta$ is the standard deviation on the bearing noise in radians,
$\sigma_\textrm{r}$ is the standard deviation on the range noise in meters, and
$\delta$ is a small value included to ensure \eqref{eqn:noise_cov_SE2} remains
positive definite.  Note that the notation $\delta \mbs{\xi}^s_m$ describes a
perturbation about point $s$ resolved in frame $\rframe{m}$.  The covariance on
the range-bearing sensor \textit{measurement} is therefore defined by
$\mbs{\Sigma}^p_m$, which is related to sensor covariance $\mbs{\Sigma}^s_m$ via
the adjoint matrix, 
\begin{equation}
	\delta \mbs{\xi}^p_m = \Adj \bigl( \mbf{T}^{sp}_{mm} \bigr) \delta \mbs{\xi}^s_m.
    \label{eqn:rangebearingSE2cov}
\end{equation}
Note here that ${\delta \mbs{\xi}^s_m}$ indicates a perturbation about point $s$ while ${\delta \mbs{\xi}^p_m}$ indicates a perturbation about point $p$, with both perturbations resolved in the measurement-aligned frame $\rframe{m}$.  Hence ${\mbf{T}^{sp}_{mm} = \mbf{T}^{sp}_{mm} \big( \eye, \mbf{r}^{sp}_m \big)}$.  The $SE(2)$ RB measurement model is defined by \eqref{eqn:rangebearingSE2mean} and
\eqref{eqn:rangebearingSE2cov},
\begin{subequations}
    \begin{align}
        \mbf{T}^{ps}_{\ell m} &= \mbfbar{T}^{ps}_{\ell m} \Exp \big( \delta \mbs{\xi}^p_m \big), \\
        \mbs{\Sigma}^p_m  &= \expect{\, \delta \mbs{\xi}^p_m \, (\delta \mbs{\xi}^p_m)^\trans \,} = \Adj \bigl( \mbf{T}^{sp}_{mm} \bigr) \mbs{\Sigma}^s_m \Adj \bigl( \mbf{T}^{sp}_{mm} \bigr)^\trans.
    \end{align}
    \label{eqn:rangebearingSE2model}%
\end{subequations}
An illustration of this noise model is shown in \Cref{fig:range_bearing_matlab}.
The DCM $\mbf{C}_{\ell m}$ may be found through Gram-Schmidt orthogonalization.
Defining ${\mbf{m}^1_\ell = \mbf{r}^{ps}_\ell \, / \, \| \mbf{r}^{ps}_\ell \|}$
to lie along the range direction, $\mbf{m}^2_\ell$ is aligned with $\mbs{\ell}^2_\ell
= \begin{matrix} [ \, 0 & 1 \, ] \end{matrix}^\trans$ via 
\begin{subequations}
    \begin{align}
        \mbf{u}^2_\ell =& \ \mbs{\ell}^2_\ell - ((\mbf{m}^1_\ell)^\trans \mbs{\ell}^2_\ell) \cdot \mbf{m}^1_\ell, \\
        \mbf{m}^2_\ell =& \ \mbf{u}^2_\ell \, / \, \| \mbf{u}^2_\ell \|,
    \end{align}
\end{subequations}
with the DCM given by 
\begin{equation}
    \mbf{C}_{\ell m} = \begin{bmatrix}
        \mbf{m}^1_\ell & \mbf{m}^2_\ell
    \end{bmatrix} \in SO(2).
\end{equation}

The generalized $SE(3)$ RAE measurement model takes the same form as the $SE(2)$
RB model given in \eqref{eqn:rangebearingSE2model}, but with the poses,
perturbations, and adjoints defined on $SE(3)$ (see, respectively,
\eqref{eqn:elementSE3} and \eqref{eqn:adjSE3}).  The pose measurement becomes
${\mbf{T}^{ps}_{\ell m} \in SE(3)}$, with ${\mbf{r}^{ps}_\ell(r) = \mbf{C}_{\ell
m} \mbf{r}^{ps}_m}$, ${\mbf{r}^{ps}_m =
\begin{matrix} [ \, r & 0 & 0 \, ] \end{matrix}^\trans}$.  The DCM
$\mbf{C}_{\ell m}$ is still obtained through orthogonalization, with
${\mbs{\ell}^2_\ell = \begin{matrix} [ \, 0 & 1 & 0 \, ] \end{matrix}^\trans}$
and ${\mbf{m}^3_\ell = (\mbf{m}^1_\ell)^\times \mbf{m}^2_\ell}$, leading to 
\begin{equation}
    \mbf{C}_{\ell m} = \begin{bmatrix}
        \mbf{m}^1_\ell & \mbf{m}^2_\ell & \mbf{m}^3_\ell
    \end{bmatrix} \in SO(3).
    \label{eqn:ClmSO3}
\end{equation}
The covariance on the RAE sensor is now 
\begin{equation}
    \mbs{\Sigma}^s_m = \diag \bigl( \sigma^2_\theta, \sigma^2_\phi, \delta^2, \sigma^2_\textrm{r}, \delta^2, \delta^2 \bigr),
    \label{eqn:noise_cov_SE3}
\end{equation}
with ${\sigma_\textrm{r}, \sigma_\phi, \sigma_\theta}$ the standard
deviations on the range, azimuth, and elevation noise, respectively, with units
of \SI{}{\meter}, \SI{}{\radian}, \SI{}{\radian}, and again where $\delta$ is a
small value included to ensure \eqref{eqn:noise_cov_SE3} remains positive
definite.  An illustration of this noise model is shown in
\Cref{fig:rae_matlab}.

% -----------------------------------------------------------
% -----------------------------------------------------------
\subsection{Part II: Including Sensor-to-Vehicle Extrinsic Uncertainty}
\label{sec:extrinsics}
Sensor-to-vehicle extrinsic calibration, whereby the pose of a sensor is
estimated relative to the vehicle, is a critical aspect of robotics.  Sensor
extrinsics must often by calibrated in observability-constrained environments,
for example on ground vehicles \cite{Lv2022a} or rotationally stable subsea
vehicles \cite{Hitchcox2024}, resulting in calibration parameters with varying
degrees of certainty.  This uncertainty should be accounted for when resolving
sensor measurements in the vehicle reference frame.  

This is easily accomplished using the developed $SE(3)$ noise model.  With the
sensor-to-vehicle extrinsics described by ${\mbf{T}^{sz}_{b\ell} \in SE(3)}$,
the RAE measurement of point $p$ is expressed relative to the vehicle as 
\begin{equation}
    \mbf{T}^{pz}_{bm} = \mbf{T}^{sz}_{b\ell} \,\mbf{T}^{ps}_{\ell m}.
\end{equation}
Applying a right perturbation yields
\begin{align*}
    \mbf{T}^{pz}_{bm} &= \mbfbar{T}^{sz}_{b\ell} \Exp(\delta \mbs{\xi}^s_\ell) \, \mbfbar{T}^{ps}_{\ell m} \Exp(\delta \mbs{\xi}^p_m) \\ 
    &\approx \mbfbar{T}^{pz}_{bm} \Exp \bigl( \Adj \bigl( \bigl( \mbfbar{T}^{ps}_{\ell m} \bigr)\inv \bigr) \delta \mbs{\xi}^s_\ell + \delta \mbs{\xi}^p_m \bigr).
    \numberthis
    \label{eqn:ext_1}
\end{align*}
Assuming the uncertainty in the sensor extrinsics and sensor measurement in
\eqref{eqn:ext_1} are uncorrelated, the vehicle-frame sensor measurement becomes 
\begin{subequations}
    \begin{align}
        \mbf{T}^{pz}_{bm} &= \mbfbar{T}^{pz}_{bm} \Exp( \delta \mbs{\eta}^p_m ), \\
        \mbs{\Xi}^p_m &= \expect{ \, \delta \mbs{\eta}^p_m \, (\delta \mbs{\eta}^p_m)^\trans \, } \nonumber \\
        &= \Adj \bigl( \bigl( \mbfbar{T}^{ps}_{\ell m} \bigr) \inv \bigr) \mbs{\Sigma}^s_\ell \Adj \bigl( \bigl( \mbfbar{T}^{ps}_{\ell m} \bigr) \inv \bigr)^\trans + \mbs{\Sigma}^p_m,
        \label{eqn:noise_cov_bodyframe}
    \end{align}
    \label{eqn:noise_meas_model_SE3_rframe_b}%
\end{subequations}
where the symbol $\mbs{\Xi}^p_m$ has been used to disambiguate from
$\mbs{\Sigma}^p_m$ in \eqref{eqn:rangebearingSE2model}, and where 
\begin{equation}
    \mbs{\Sigma}^s_\ell = \expect{ \, \delta \mbs{\xi}^s_\ell \, (\delta \mbs{\xi}^s_\ell)^\trans \, }
    \label{eqn:noise_extrinsics}
\end{equation}
represents the covariance on the sensor-to-vehicle extrinsic estimate
$\mbf{T}^{sz}_{b \ell}$.

% -----------------------------------------------------------
% -----------------------------------------------------------
\subsection{Part III: Uncertainty in the Vehicle Motion Estimate}
\label{sec:vehicle_motion}
Consider \Cref{fig:motivating_example}, in which a submap is constructed from
individual range-bearing measurements of a wall.  Vehicle pose
$\mbf{T}^{z_1w}_{ab_1}$, on the left, has been selected as the ``central''
submap pose for the purpose of scan alignment.  The submap-level noise model for
point $p_2$, measured from sensor location $s_2$, clearly needs to incorporate
the odometry uncertainty accumulated between the central submap pose and the
vehicle pose $\mbf{T}^{z_2w}_{ab_2}$ from which the measurement of $p_2$ was
recorded.  

To simplify the notation, let $\mbf{T}^{\bar{z}w}_{a\bar{b}}(\mbf{C}_{a\bar{b}},
\mbf{r}^{\bar{z}w}_a)$ denote the central submap pose, and let
$\mbf{T}^{zw}_{ab}$ indicate the pose at which some point $p$ is measured.
Generalizing to $SE(3)$, the RAE measurement of point $p$ is expressed relative
to the central submap pose as 
\begin{equation}
    \mbf{T}^{p\bar{z}}_{\bar{b}m} = \mbf{T}^{z\bar{z}}_{\bar{b}b} \, \mbf{T}^{pz}_{bm}.  
\end{equation}
Applying a right perturbation yields
\begin{align*}
    \mbf{T}^{p\bar{z}}_{\bar{b}m} &= \mbfbar{T}^{z\bar{z}}_{\bar{b}b} \Exp(\delta \mbs{\xi}^{z}_{b}) \, \mbfbar{T}^{pz}_{bm} \Exp(\delta \mbs{\eta}^{p}_{m}) \\ 
    &\approx \mbfbar{T}^{p\bar{z}}_{\bar{b}m} \Exp \bigl( \Adj \bigl( \bigl( \mbfbar{T}^{pz}_{bm} \bigr)\inv \bigr) \delta \mbs{\xi}^z_b + \delta \mbs{\eta}^p_m \bigr).
    \numberthis
    \label{eqn:ext_2}
\end{align*}
Assuming the uncertainty in the odometry estimate and the measurement
uncertainty are uncorrelated, the submap-level measurement of point $p$
becomes 
\begin{subequations}
    \begin{align}
        \mbf{T}^{p\bar{z}}_{\bar{b}m} &= \mbfbar{T}^{p\bar{z}}_{\bar{b}m} \Exp( \delta \mbs{\zeta}^p_m ), \\
        \mbs{\Gamma}^p_m &= \expect{ \, \delta \mbs{\zeta}^p_m \, (\delta \mbs{\zeta}^p_m)^\trans \, } \nonumber \\
        &= \Adj \bigl( \bigl( \mbfbar{T}^{pz}_{bm} \bigr)\inv \bigr) \mbs{\Sigma}^z_b \Adj \bigl( \bigl( \mbfbar{T}^{pz}_{bm} \bigr)\inv \bigr)^\trans + \mbs{\Xi}^p_m,
        \label{eqn:noise_cov_submap}
    \end{align}
    \label{eqn:noise_meas_model_SE3_submap}%
\end{subequations}
in which $\mbs{\Xi}^{p}_{m}$ is the vehicle-frame measurement noise of $p_2$
given by \eqref{eqn:noise_cov_bodyframe} and where 
\begin{equation}
    \mbs{\Sigma}^z_b = \expect{ \, \delta \mbs{\xi}^z_b \, (\delta \mbs{\xi}^z_b)^\trans \, }
    \label{eqn:relative_pose}
\end{equation}
is the uncertainty in the relative pose estimate $\mbf{T}^{z\bar{z}}_{\bar{b}b}$
(see \cite[(51)]{Mangelson2019}) caused by noisy and potentially biased odometry
measurements between the central vehicle position $\bar{z}$ and $z$.

% -----------------------------------------------------------
% -----------------------------------------------------------
\subsection{Summary: A Submap-Level Noise Model for RAE Sensors}
\label{sec:summary}
Incorporating the sensor-level RAE noise model \eqref{eqn:rangebearingSE2model}
and the vehicle-level noise model \eqref{eqn:noise_meas_model_SE3_rframe_b} into
the submap-level noise model \eqref{eqn:noise_meas_model_SE3_submap} and
simplifying yields the final compound model, 
\begin{subequations}
    \begin{align}
        \mbfbar{T}^{p\bar{z}}_{\bar{b}m} &= \mbfbar{T}^{z\bar{z}}_{\bar{b}b} \, \mbfbar{T}^{sz}_{b\ell} \, \mbfbar{T}^{ps}_{\ell m}, \\
        \mbs{\Gamma}^p_m &= \expect{ \, \delta \mbs{\zeta}^p_m \, (\delta \mbs{\zeta}^p_m)^\trans \, } \nonumber \\
        \begin{split}
            &= \Adj( \mbfbar{T}^{zp}_{mb} ) \mbs{\Sigma}^z_b \Adj( \mbfbar{T}^{zp}_{mb} )^\trans + \Adj( \mbfbar{T}^{sp}_{m\ell} ) \mbs{\Sigma}^s_\ell \Adj( \mbfbar{T}^{sp}_{m\ell} )^\trans \\
                & \qquad + \Adj( \mbfbar{T}^{sp}_{mm} ) \mbs{\Sigma}^s_m \Adj( \mbf{T}^{sp}_{mm} )^\trans.
        \end{split}
        \label{eqn:noise_meas_model_SE3_full_cov}%
    \end{align}
    \label{eqn:noise_meas_model_SE3_full}%
\end{subequations}
The three covariance matrices that appear in
\eqref{eqn:noise_meas_model_SE3_full_cov} are the RAE sensor covariance
\eqref{eqn:noise_cov_SE3}, the sensor-to-vehicle extrinsic uncertainty
\eqref{eqn:noise_extrinsics}, and the uncertainty in the relative pose estimate
\eqref{eqn:relative_pose}.  The nontrivial ingredients needed to construct the
various adjoints throughout \eqref{eqn:noise_meas_model_SE3_full_cov} are 
\begin{subequations}
    \begin{align}
        \mbf{r}^{sp}_m &= \begin{bmatrix}
            -r & 0 & 0
        \end{bmatrix}^\trans, \\
        \mbf{C}_{m \ell} &= \mbf{C}_{\ell m}^\trans, \\
        \mbf{C}_{mb} &= \mbf{C}_{m \ell} \mbf{C}_{b\ell}^\trans, \\
        \mbf{r}^{zp}_m &= -\mbf{C}_{mb} \mbf{r}^{sz}_b + \mbf{r}^{sp}_m.
    \end{align}
\end{subequations}
These expressions are directly constructed from the range measurement $r$, the
DCM $\mbf{C}_{\ell m}$ \eqref{eqn:ClmSO3}, and the sensor-to-vehicle extrinsics
$\mbf{T}^{sz}_{b \ell}(\mbf{C}_{b \ell}, \mbf{r}^{sz}_b)$.  The compound $SE(3)$
noise model given by \eqref{eqn:noise_meas_model_SE3_full} therefore represents
a simple, inexpensive method of incorporating multiple sources of uncertainty
into a scan matching problem.  

This compound noise model will be beneficial for robust data association
applications, for example measurement-to-landmark association \cite{Retan2022}
and robust point-cloud alignment \cite{Xu2024}.  Conventional point-cloud
alignment algorithms typically model point measurements as vector elements,
${\mbf{r}^{p\bar{z}}_{\bar{b}} \in \rnums^3}$, ${\mbf{r}^{p\bar{z}}_{\bar{b}} =
\mbfbar{r}^{p\bar{z}}_{\bar{b}} + \delta \mbf{r}_{\bar{b}}}$.  The covariance on
this form of the measurement is extracted from the $SE(3)$ covariance
$\mbs{\Gamma}^p_m$ as
\begin{subequations}
    \begin{align}
        \mbs{\Gamma}^p_m &= \begin{bmatrix}
            \mbs{\Gamma}_m^{\phi \phi} & \mbs{\Gamma}_m^{\phi \rho} \\ 
            \mbs{\Gamma}_m^{\rho \phi} & \mbs{\Gamma}_m^{\rho \rho}
        \end{bmatrix}, \\
        \mbs{\Sigma}_{\rnums^3} &= \expect{ \delta \mbf{r}_{\bar{b}} (\delta \mbf{r}_{\bar{b}})^\trans } = \mbfbar{C}_{\bar{b}m} \mbs{\Gamma}_m^{\rho \rho} \, \mbfbar{C}_{\bar{b}m}^\trans,
    \end{align}
\end{subequations}
with ${\mbf{C}_{\bar{b}m} = \mbf{C}_{\bar{b}b} \mbf{C}_{mb}^\trans}$.  Though
projected back into $\rnums^3$, the resulting covariance will incorporate all
angular uncertainty, adjoint effects, and odometry uncertainty represented in
\eqref{eqn:noise_meas_model_SE3_full_cov}, which includes the uncertainty from
noisy and biased interoceptive measurements.  The full noise model
\eqref{eqn:noise_meas_model_SE3_full} could also be used directly for
applications involving a residual defined on $SE(3)$.

% -----------------------------------------------------------
% -----------------------------------------------------------
\section{Results}
\label{sec:results}

This section examines the behaviour of the developed noise models in two quantitative applications.  The first application revisits the motivating example from \Cref{sec:motivatingexample}, and illustrates how noisy odometry can affect the resulting $SE(2)$ submap noise profile.  The second application constructs an $SE(3)$ noise profile using a white-noise-on-acceleration motion prior and real data from an underwater laser scanner.

% -----------------------------------------------------------
% -----------------------------------------------------------
\subsection{Simulation Results: Motivating Example}
\label{sec:results_sim}

Consider again the motivating example illustrated in \Cref{fig:motivating_example}, in which a vehicle moves from left to right whilst recording range-bearing measurements of a wall.  For example, these measurements could be recorded using lidar, a ``push-broom'' laser scanner \cite{Baldwin2012}, radar \cite{Xu2024}, or other sensor for which a range-bearing model is appropriate.  As before, the individual measurements must be combined into a single submap in order to facilitate scan matching.  Quantitatively, how will the uncertainty in the range-bearing measurements, sensor-to-vehicle pose estimate, and vehicle odometry combine to determine the noise profile of the submap?

\begin{figure*}
	% preliminary
	\sbox\subfigbox{%
	  \resizebox{\dimexpr0.96\textwidth-1em}{!}{%
		\includegraphics[height=3cm]{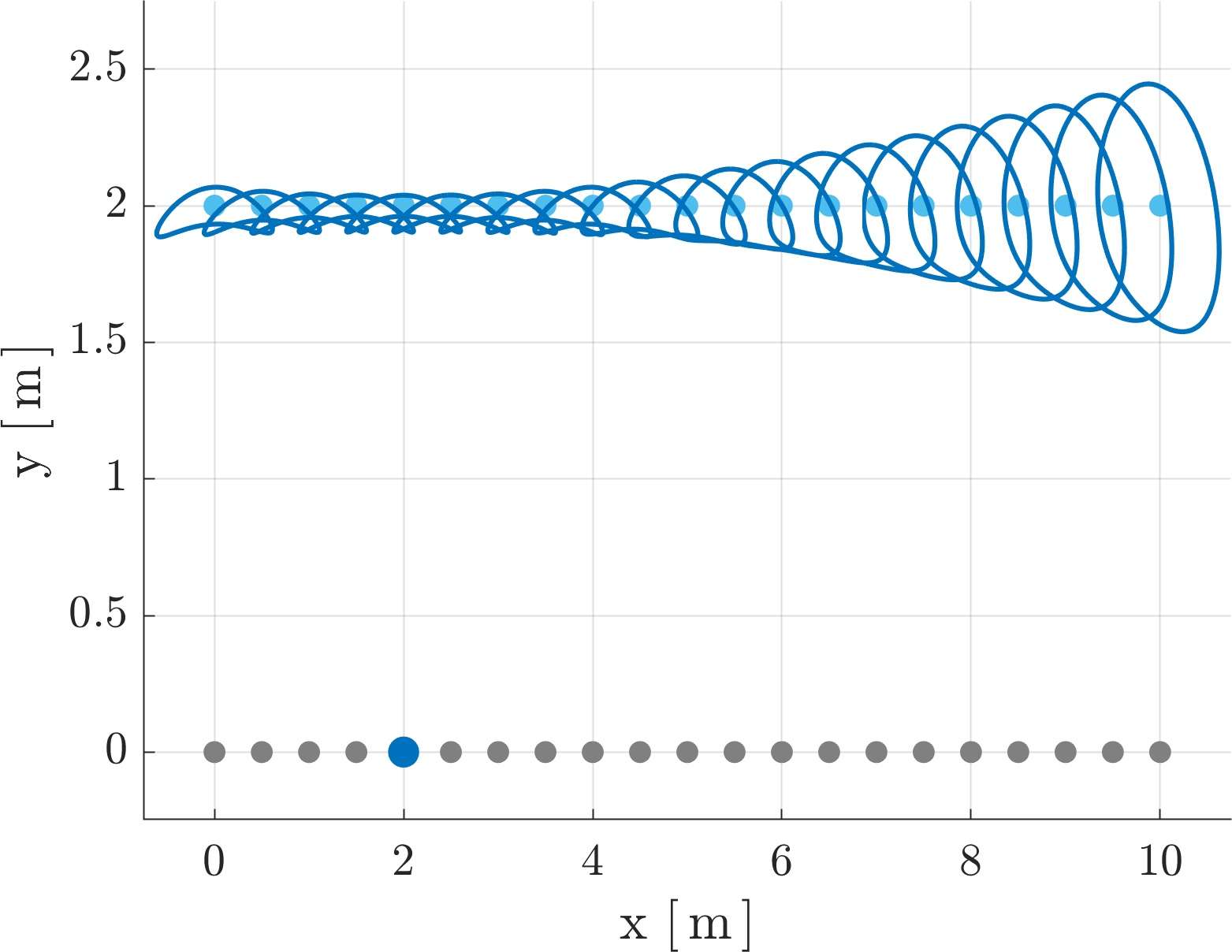}%
        \includegraphics[height=3cm]{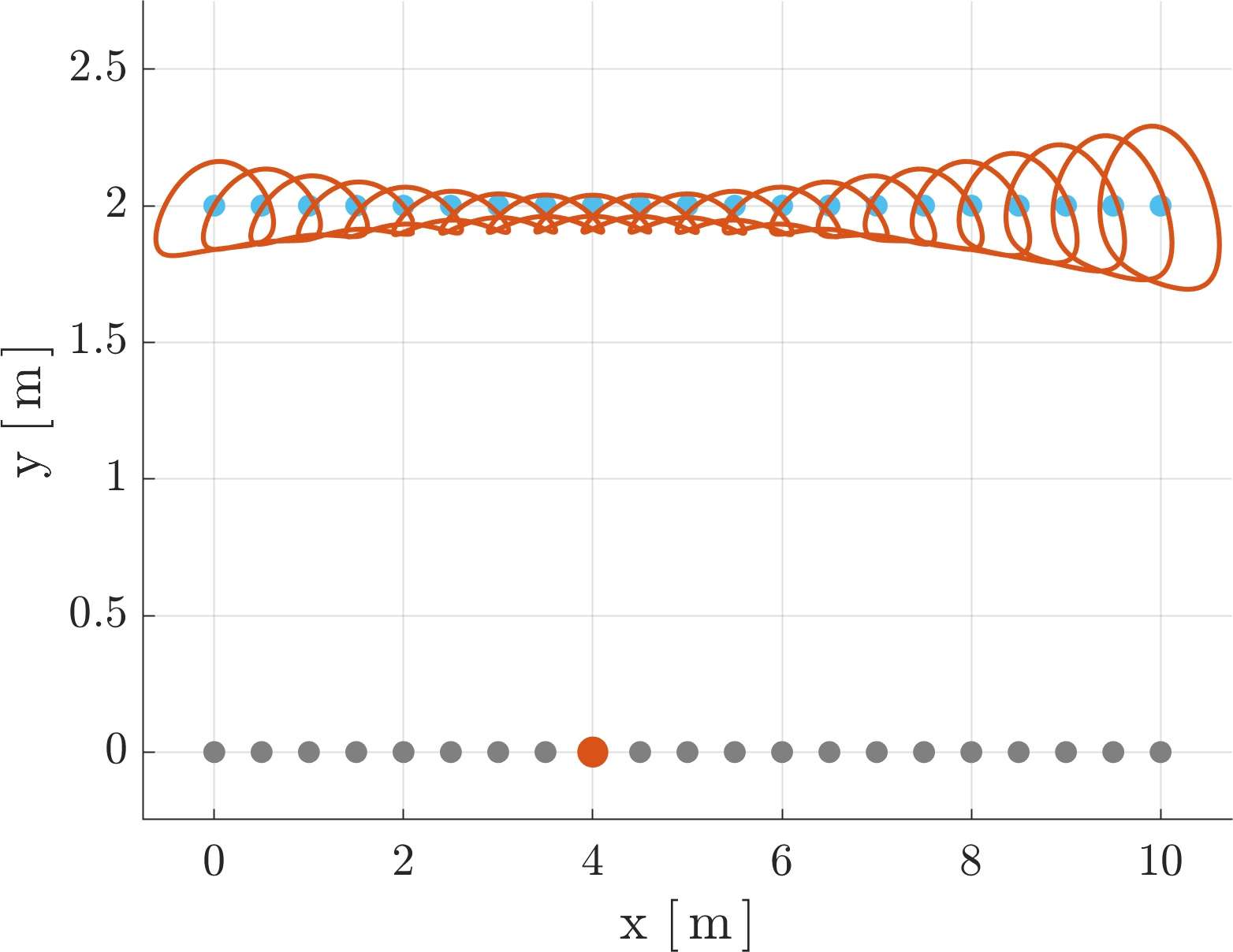}%
		\includegraphics[height=3cm]{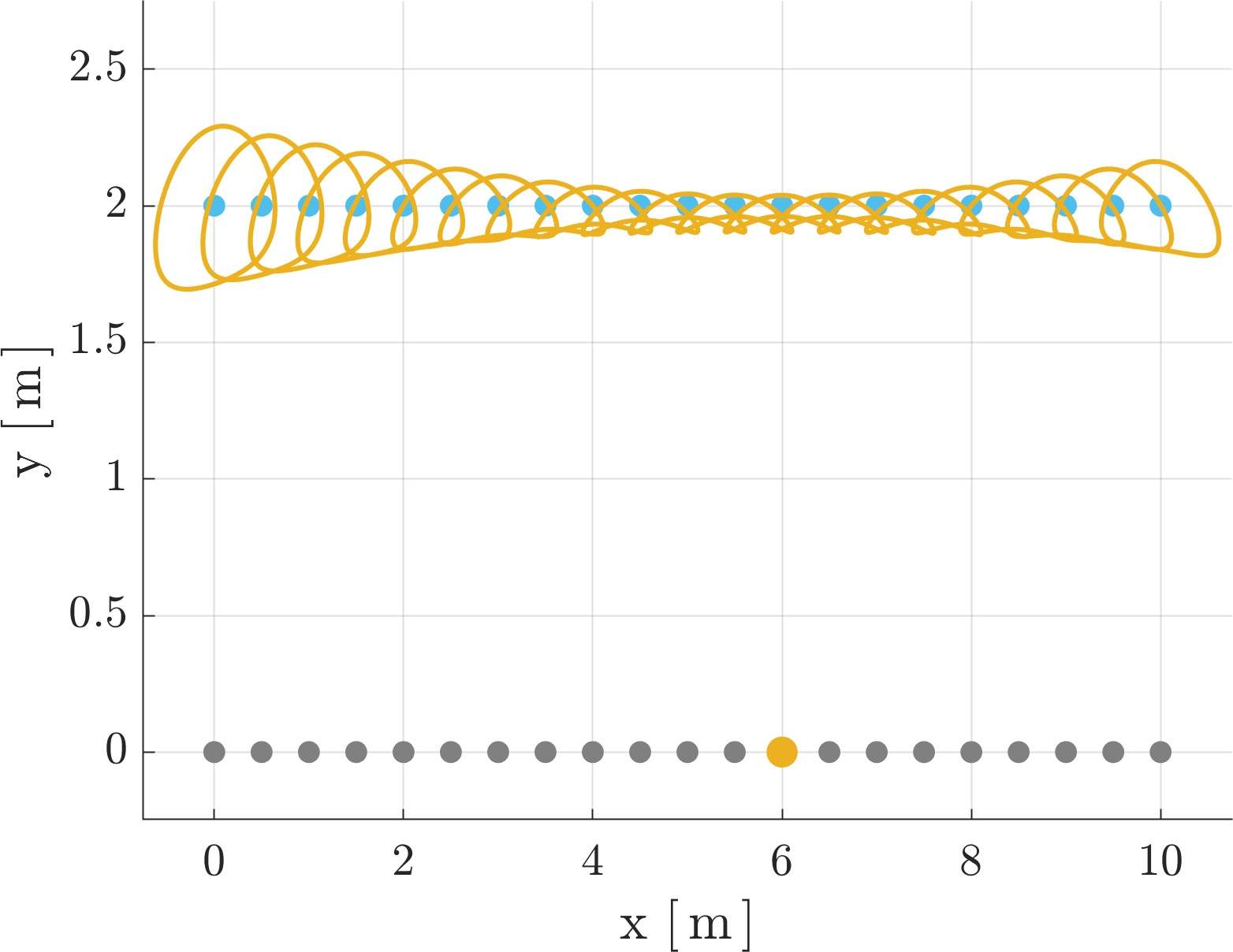}%
		\includegraphics[height=3cm]{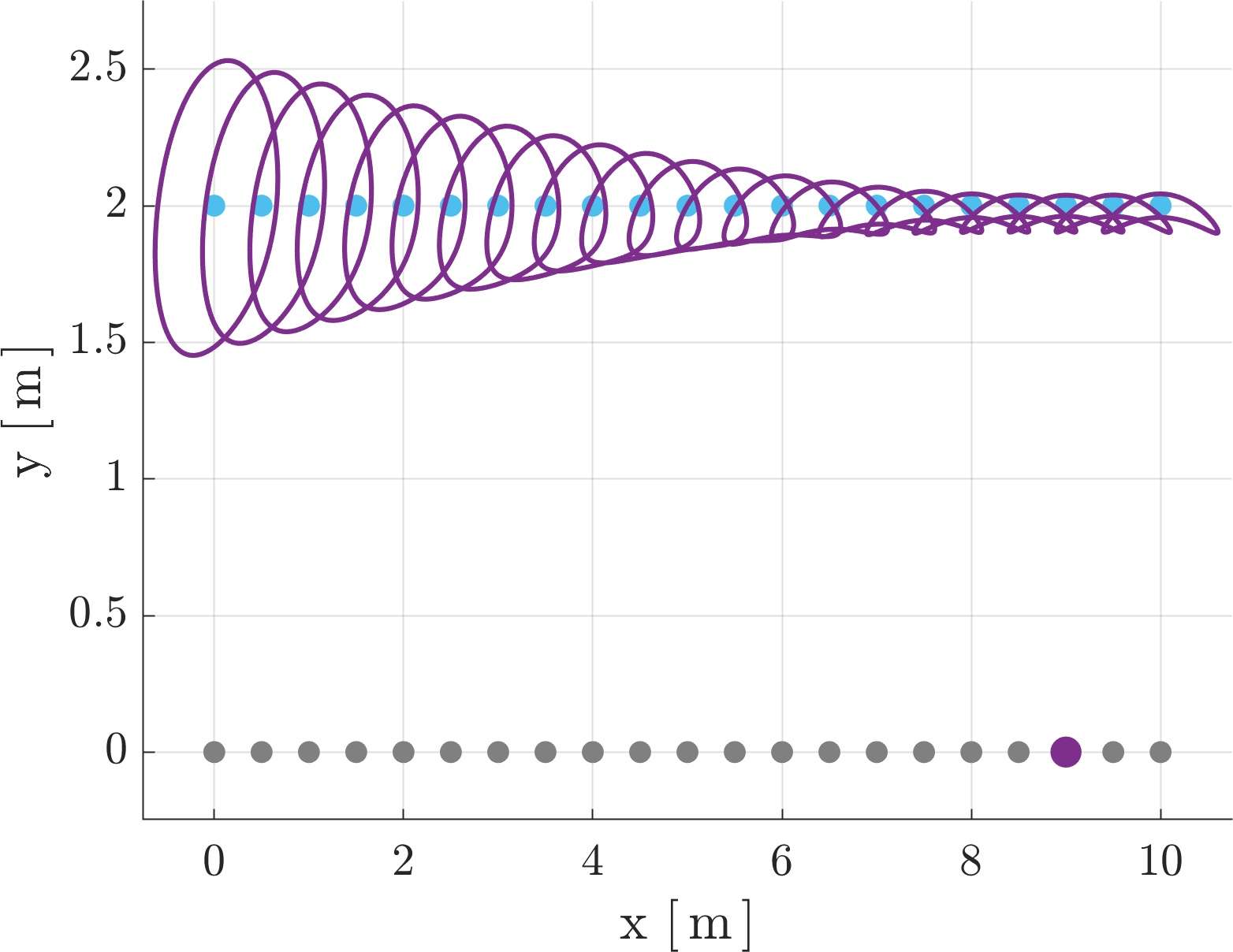}%
	  }%
	}
    \setlength{\subfigheight}{\ht\subfigbox}
	\centering
    % the figure
	\subcaptionbox{Central pose 5 \label{fig:cov_pt_5}}{%
	  \includegraphics[height=\subfigheight]{figs/se2_covariance/compound_noise_central_pose_5.jpg}
	}
    \subcaptionbox{Central pose 9 \label{fig:cov_pt_9}}{%
        \includegraphics[height=\subfigheight]{figs/se2_covariance/compound_noise_central_pose_9.jpg}
    }
    \subcaptionbox{Central pose 13 \label{fig:cov_pt_13}}{%
	  \includegraphics[height=\subfigheight]{figs/se2_covariance/compound_noise_central_pose_13.jpg}
	}
    \subcaptionbox{Central pose 19 \label{fig:cov_pt_19}}{%
	  \includegraphics[height=\subfigheight]{figs/se2_covariance/compound_noise_central_pose_19.jpg}
	}
    \par \bigskip
    \subcaptionbox{Meas. 5, superimposed \label{fig:cov2_pt_5}}{%
        \includegraphics[height=\subfigheight]{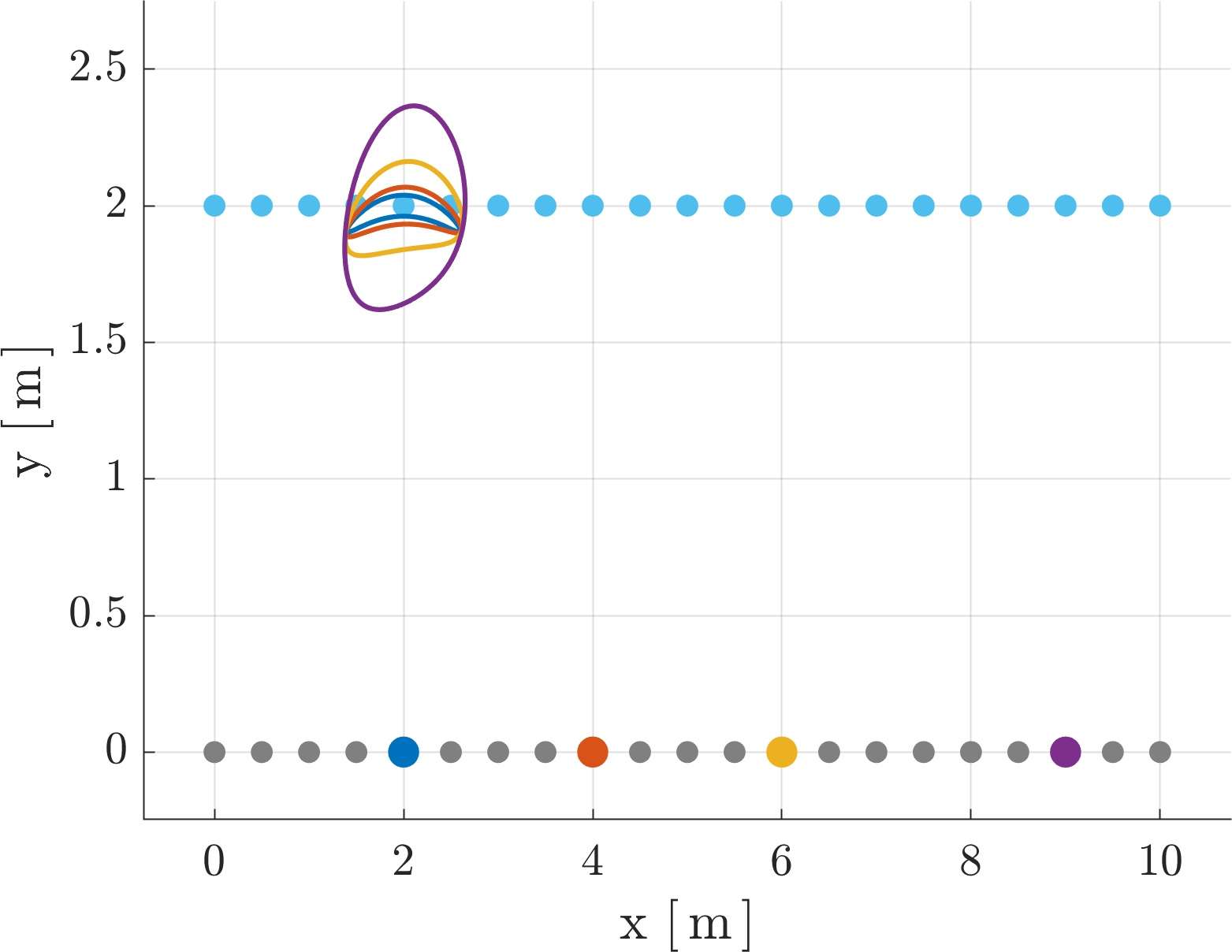}
    }
    \subcaptionbox{Meas. 9, superimposed \label{fig:cov2_pt_9}}{%
        \includegraphics[height=\subfigheight]{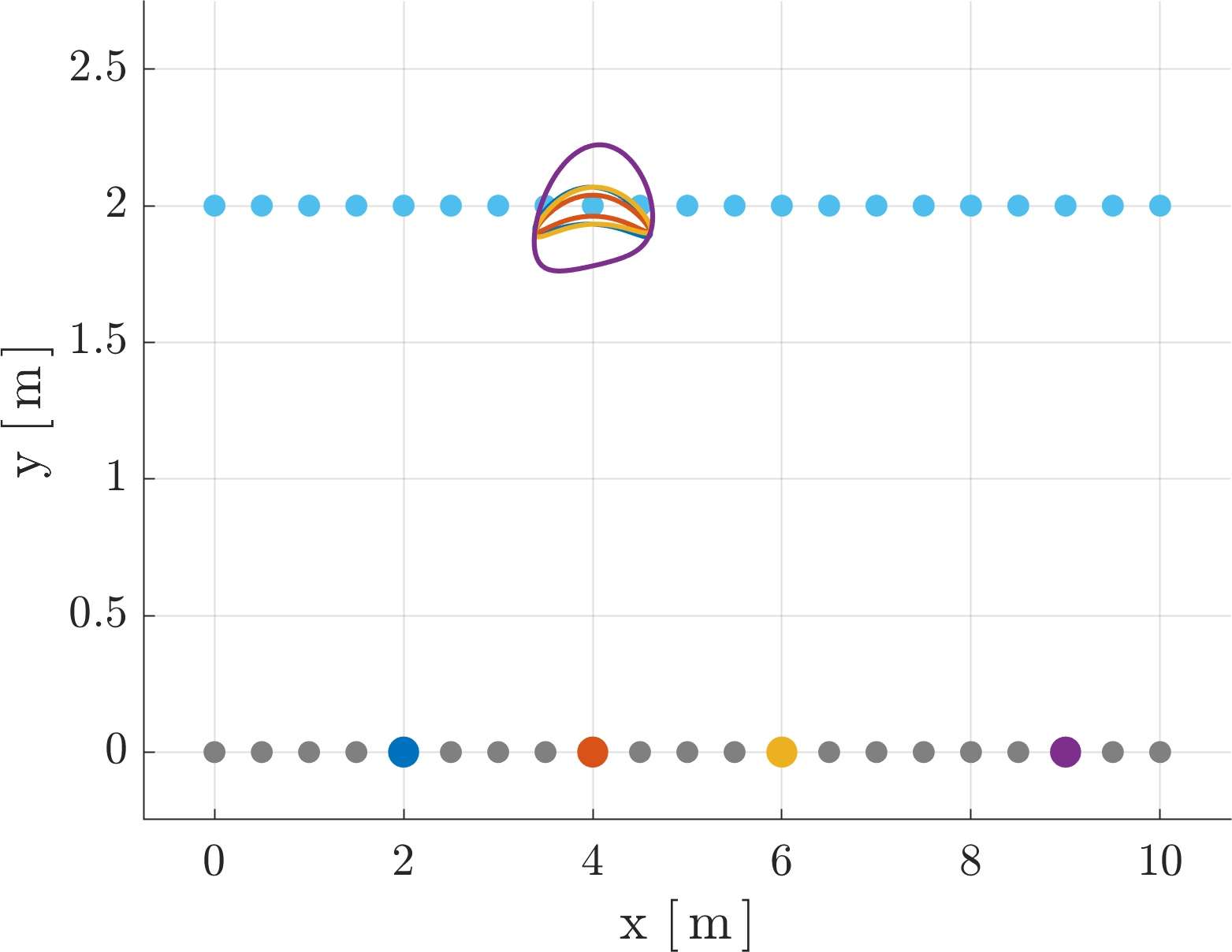}
    }
    \subcaptionbox{Meas. 13, superimposed \label{fig:cov2_pt_13}}{%
        \includegraphics[height=\subfigheight]{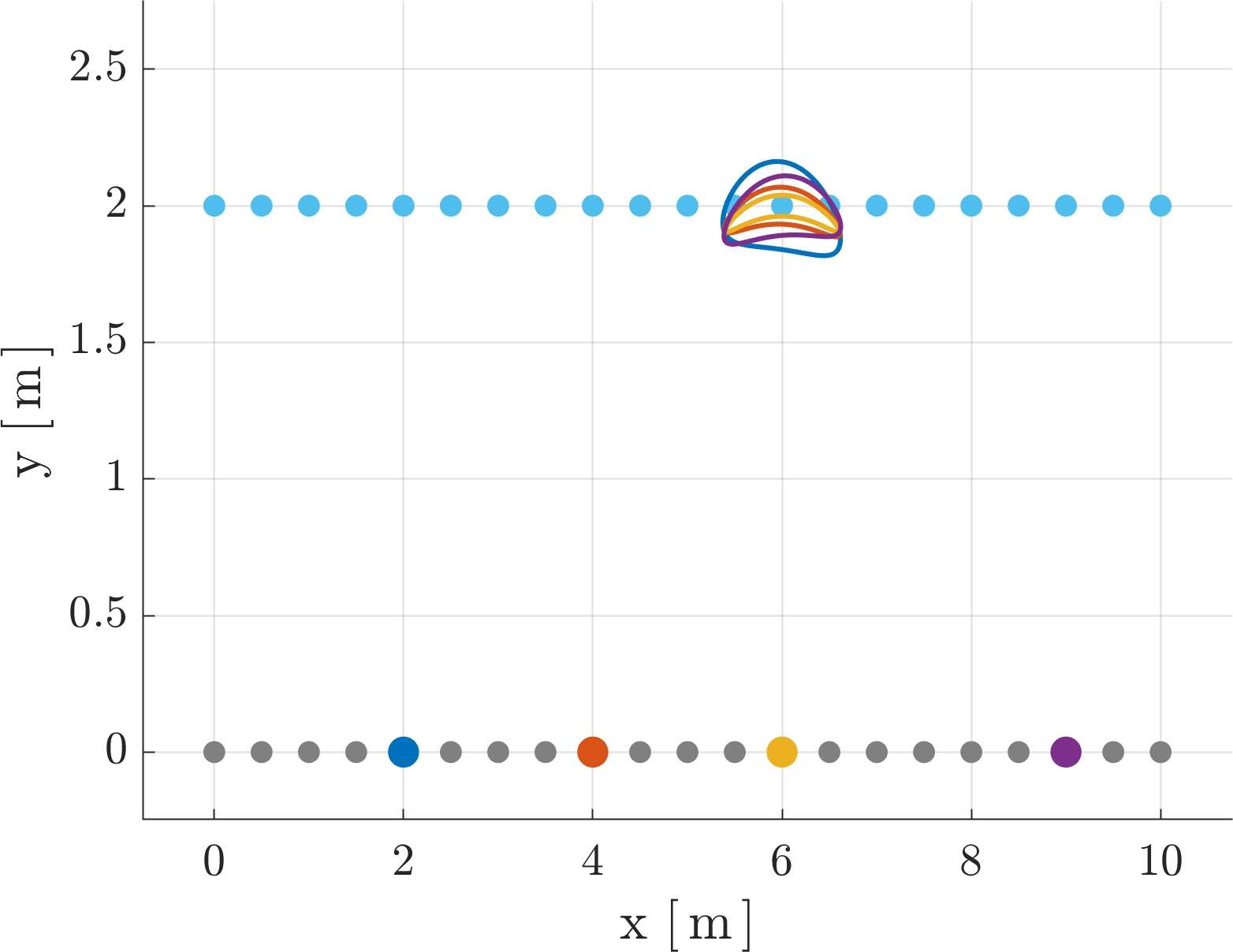}
    }
    \subcaptionbox{Meas. 19, superimposed \label{fig:cov2_pt_19}}{%
        \includegraphics[height=\subfigheight]{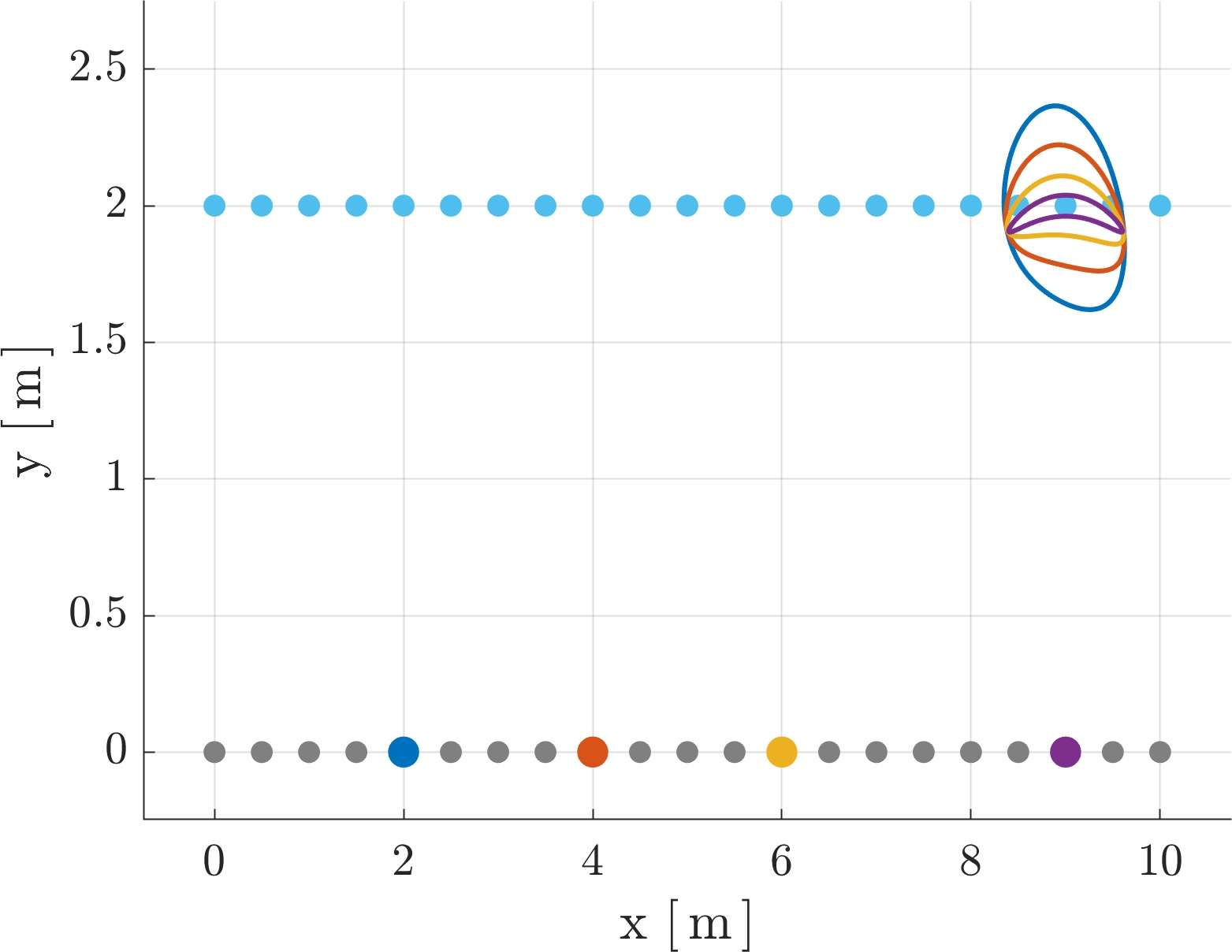}
    }
    \caption{Uncertainty envelopes for the submap-level noise model \eqref{eqn:noise_meas_model_SE3_full_cov}.  A vehicle moves from left to right, periodically recording range-bearing measurements of a wall at ${y = \SI{2}{\meter}}$.  When the individual measurements are combined into a submap about a single ``central'' pose, the resulting uncertainty envelope should incorporate measurement uncertainty, sensor-to-vehicle extrinsic uncertainty, and odometry uncertainty.  The top row shows \SI{99.73}{\percent} uncertainty envelopes for different selections of the central submap pose (highlighted), while the bottom row shows the four sets of envelopes superimposed on selected measurements.}
  \label{fig:results_sim}
\end{figure*}

\Cref{fig:results_sim} summarizes the results from a small simulation illustrating these effects.  A vehicle moves forward at ${\SI{1}{\meter\per\second}}$ from ${x = \SI{0}{\meter}}$ to ${x = \SI{10}{\meter}}$ across the bottom of each figure, recording at ${\SI{0.5}{\meter}}$ intervals a single range-bearing measurement of a wall located at ${y = \SI{2}{\meter}}$.  The vehicle positions are illustrated by grey dots, while the measurements are illustrated by cyan dots.  The range-bearing sensor is represented using the $SE(2)$ model from \eqref{eqn:rangebearingSE2model}, with ${\sigma_{\textrm{r}} = \SI{1}{\centi\meter}}$ and ${\sigma_\theta = \SI{5}{\deg}}$.  The sensor is mounted relative to the vehicle with an  uncertainty of ${\mbs{\Sigma}^s_\ell = \diag\bigl( \sigma^2_\alpha, \sigma^2_\beta \cdot \eye \bigr) \in \rnums^{3\times 3}}$ (see \eqref{eqn:noise_extrinsics}), with ${\sigma_\alpha = \SI{1}{\deg}}$ and ${\sigma_\beta = \SI{5}{\milli\meter}}$.  

Finally, the vehicle is equipped with two interoceptive sensors.  A gyroscope measures body-frame angular velocity, 
\begin{equation}
    u^\textrm{g}_b = \bar{u}^\textrm{g}_b + w^\textrm{g}_b,
\end{equation}
with ${w^\textrm{g}_b \sim \mathcal{N} \bigl( 0, \sigma^2_\textrm{g} \bigr)}$, and a wheel encoder measures forward body velocity, 
\begin{equation}
    u^\textrm{w}_b = \bar{u}^\textrm{w}_b + w^\textrm{w}_b, 
\end{equation}
with ${w^\textrm{w}_b \sim \mathcal{N} \bigl( 0, \sigma^2_\textrm{w} \bigr)}$.  For the simulation, the power spectral density (PSD) on the noise associated with each sensor is taken to be ${\sigma^2_\textrm{g} = \SI{1e-4}{\radian\squared\second\tothe{-1}}}$ and ${\sigma^2_\textrm{w} = \SI{1e-4}{\meter\squared\second\tothe{-1}}}$, respectively.  \Cref{tab:noise_motivating_example_table} summarizes the simulation parameters.  The range measurement uncertainty and sensor-to-vehicle mounting uncertainty approximate real-world values, while the bearing measurement uncertainty and gyro PSD have generally been inflated for illustrative effect.  The wheel odometry PSD is characteristic of a low-grade wheel encoder.

\begin{table}
    \centering
    \caption{Parameters for $SE(2)$ compound noise experiment}
    \renewcommand{\arraystretch}{1.2}
    \begin{tabularx}{\columnwidth}{X|M{1cm}M{2cm}}
    \toprule
    Parameter & Symbol & Value \\
    \hline
    Std. dev. range measurement & $\sigma_{\textrm{r}}$ & \SI{1e-2}{\meter} \\
    Std. dev. bearing measurement & $\sigma_\theta$ & \SI{5}{\deg} \\
    Std. dev. extrinsic rotation & $\sigma_\alpha$ & \SI{1}{\deg} \\
    Std. dev. extrinsic translation & $\sigma_\beta$ & \SI{5e-3}{\meter} \\
    PSD gyro noise & $\sigma^2_\textrm{g}$ & \SI{1e-4}{\radian\squared\second\tothe{-1}} \\
    PSD wheel odometry noise & $\sigma^2_\textrm{w}$ & \SI{1e-4}{\meter\squared\second\tothe{-1}} \\
	\bottomrule
    \end{tabularx}
	\label{tab:noise_motivating_example_table}
\end{table}

The continuous-time $SE(2)$ vehicle kinematics are \cite{Hitchcox2023a}
\begin{equation}
    \mbfdot{T}^{zw}_{ab} = \mbf{T}^{zw}_{ab} \, \mbf{u}_b^\wedge,
\end{equation}
with ${\mbf{u}_b = \left[ u^\textrm{g}_b \ \ u^\textrm{w}_b \ \ 0 \right]^\trans}$.  Given a left-invariant error definition \cite[\textsection4.1.2]{Arsenault2019}, the associated error kinematics are 
\begin{equation}
    \delta \mbsdot{\xi} = \mbf{A} \, \delta \mbs{\xi} + \mbf{L} \, \delta \mbf{w},
\end{equation}
in which ${\mbf{A} = -\mbf{ad}(\mbfbar{u}_b)}$ and ${\mbf{L} = \left[ \eye \ \ \mbf{0} \right]^\trans \! \in \rnums^{3 \times 2}}$.  Given PSD ${\mbs{Q} = \diag \bigl( \sigma_\textrm{g}^2, \sigma_\textrm{w}^2 \bigr)}$, the discrete-time interoceptive covariance $\mbf{Q}_k$ is propagated over sampling period ${\delta t_{k-1} = t_k - t_{k-1}}$ according to \cite[\textsection4.7.2]{Farrell2008}
\begin{equation}
    \mbf{Q}_k = \int^{t_k}_{t_{k-1}} \mbf{A}(t_k - s) \mbf{L}(s) \mbc{Q}(s) \left( \mbf{A}(t_k - s) \mbf{L}(s) \right)^\trans \dee s.
    \label{eqn:Qk}
\end{equation}
This odometry uncertainty ${\mbs{\Sigma}^z_b = \mbf{Q}_k}$ (see \eqref{eqn:relative_pose}) is combined with the sensor-to-vehicle mounting uncertainty $\mbs{\Sigma}^s_\ell$ and range-bearing measurement uncertainty $\mbs{\Sigma}^s_m$ according to the submap-level noise model \eqref{eqn:noise_meas_model_SE3_full_cov}.  The top row of \Cref{fig:results_sim} shows the resulting \SI{99.73}{\percent} submap noise profile for different selections of the ``central'' submap pose $\mbf{T}^{\bar{z}w}_{a\bar{b}}$.

As the top row of \Cref{fig:results_sim} illustrates, the effects of odometry uncertainty become pronounced far away from the central submap pose.  For example, the covariance envelope directly above the central pose (blue dot) in \Cref{fig:cov_pt_5} is nearly a perfect banana shape, owing to the $SE(2)$ measurement model \eqref{eqn:rangebearingSE2model} and a small amount of uncertainty from the sensor-to-vehicle extrinsic estimate \eqref{eqn:noise_extrinsics}.  Odometry uncertainty, originating from the noisy interoceptive measurements, begins to dominate further from the central submap pose, with the furthest envelopes assuming a nearly ellipsoidal shape.  

As shown in the bottom row of \Cref{fig:results_sim}, the choice of central pose can have a large effect on the covariance envelope associated with each range-bearing measurement.  Unsurprisingly, the best option for practical applications is likely to choose a central pose near the centre of the submap, as in \Cref{fig:cov_pt_9} or \Cref{fig:cov_pt_13}, to avoid the worst effects of odometry drift.  Other selection strategies could include a central pose that is closest to the expected scan-matching location, or to choose a central pose close to information-rich areas of the submap.  Though not explored in this simulation, the effects of interoceptive bias, for example IMU measurement bias, could also easily be incorporated into the relative odometry uncertainty.

An accurate characterization of measurement uncertainty leads to more robust data association decisions \cite{Retan2022,Brossard2020}.  This simulated example shows how the developed $SE(2)$ range-bearing model, together with a careful accounting of both the sensor-to-vehicle extrinsic uncertainty and the accumulated odometry uncertainty, leads to a more accurate ``submap-level'' covariance estimate for scan-matching applications.

% -----------------------------------------------------------
% -----------------------------------------------------------
\subsection{Field Results: Wiarton Shipwreck}
\label{sec:results_field}

\begin{figure}[]
	\centering
	\includegraphics[width=0.95\columnwidth]{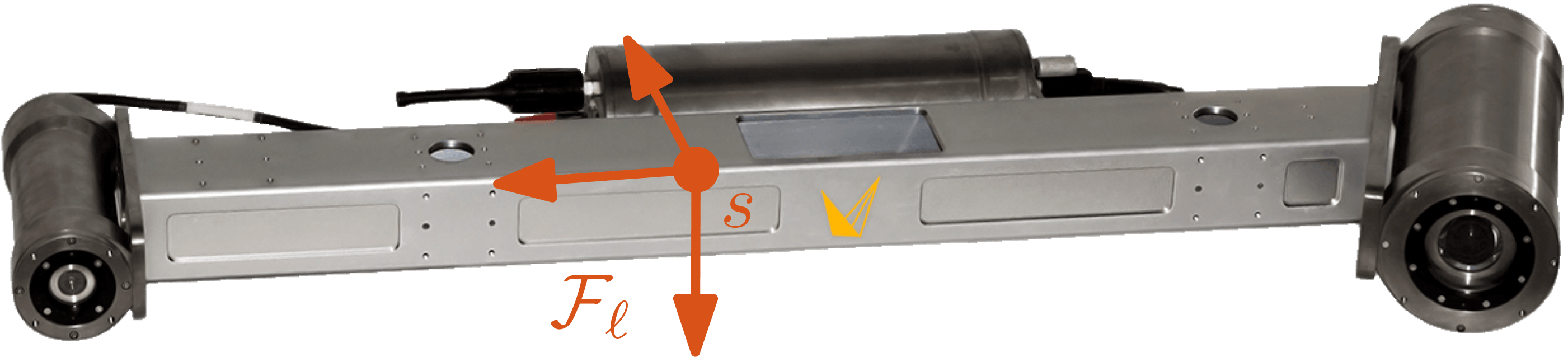}
	\caption{Voyis Insight Pro underwater laser scanner.  The baseline from laser line emitter (left) to camera (right) is one meter.}
	\label{fig:voyis_laser}
\end{figure}

\begin{figure}
	% preliminary
	\sbox\subfigbox{%
	  \resizebox{\dimexpr0.95\columnwidth-1em}{!}{%
		\includegraphics[height=3cm]{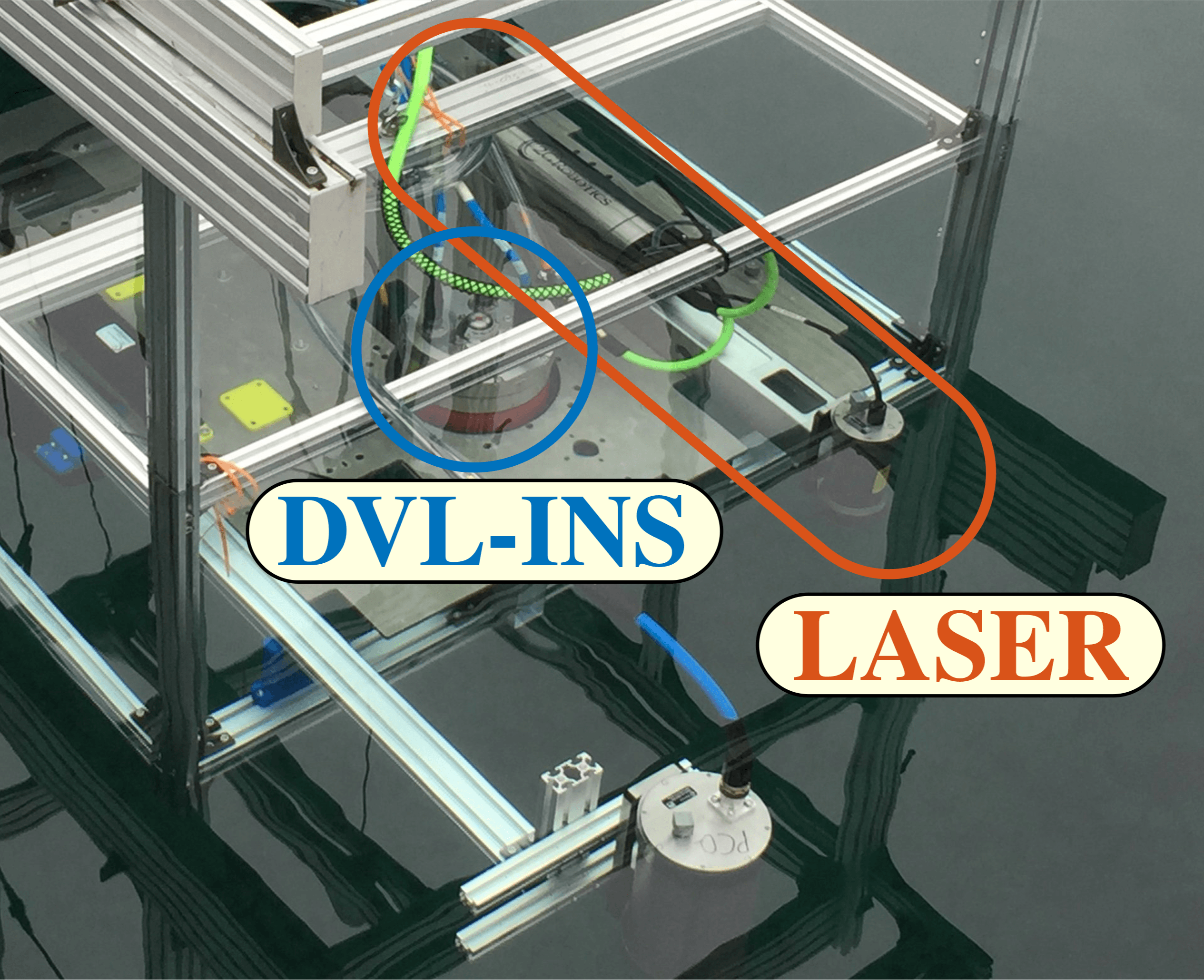}%
		\includegraphics[height=3cm]{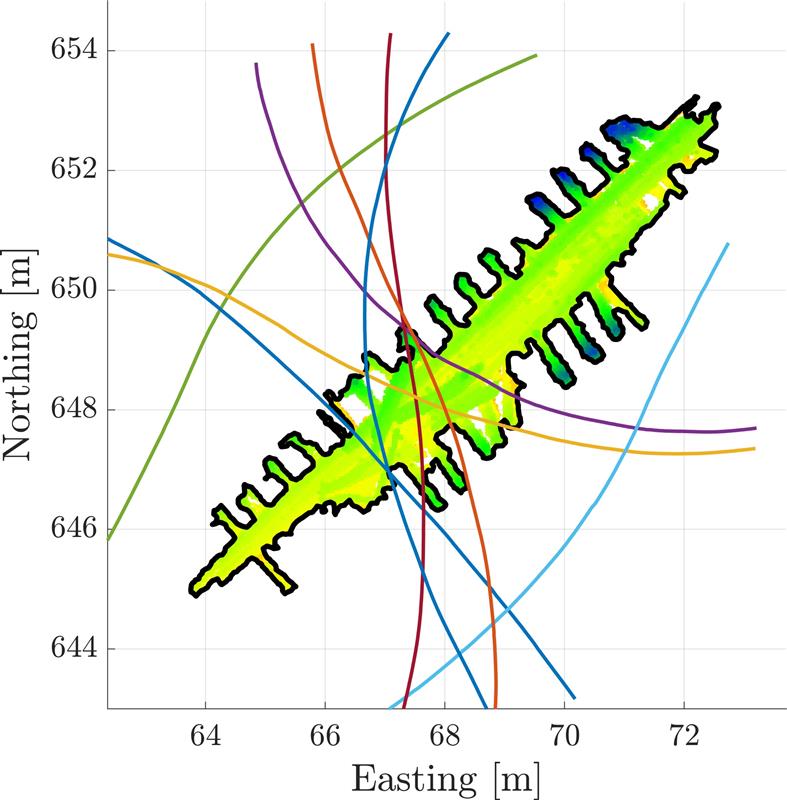}%
	  }%
	}
	\setlength{\subfigheight}{\ht\subfigbox}
	% the figure
	\centering
	\subcaptionbox{Sensor payload on surface vessel
	\label{fig:wiarton_payload}}{%
	  \includegraphics[height=\subfigheight]{figs/voyis_payload_labels_crop.png}
	}
    \hspace{3pt}
	\subcaptionbox{Eight passes over wreck \label{fig:wiarton_top_all}}{%
	  \includegraphics[height=\subfigheight]{figs/se3_covariance/wiarton_all_top.jpg}
	}
    \caption{Data collection in Wiarton, Ontario.  The sensor payload includes a DVL-INS and a Voyis laser scanner.  Laser scans were recorded during eight passes over a small shipwreck.}
    \label{fig:wiarton_overview}
\end{figure}

The full $SE(3)$ range-azimuth-elevation noise model \eqref{eqn:noise_meas_model_SE3_full_cov} is next used to visualize 3D covariance envelopes for a point-cloud submap.  Point clouds were collected using a Voyis Insight Pro underwater laser line scanner (\Cref{fig:voyis_laser}), a ``push-broom'' style sensor used commercially for recording high-resolution scans of subsea infrastructure.  The scanner emits a pulsed laser swath with a beam width of \SI{50}{\deg}, and uses an optical camera to record 2D laser profiles at a frequency of up to \SI{100}{\Hz}.  3D submaps are constructed by registering individual profiles to a vehicle trajectory estimate, provided here by a Sonardyne SPRINT-Nav Doppler velocity log-aided inertial navigation system (DVL-INS).  This sensor payload was deployed on a surface vessel and used to scan a small shipwreck in Colpoy's Bay, Wiarton, Ontario, Canada.  The sensor payload and shipwreck structure are shown in \Cref{fig:wiarton_overview}.

The laser line scanner is represented as an RAE sensor using the $SE(3)$ noise model \eqref{eqn:noise_cov_SE3}.  For illustrative purposes, the range uncertainty is taken to be ${\sigma_\textrm{r} = \SI{1}{\centi\meter}}$, the azimuth uncertainty to be ${\sigma_\phi = \SI{3}{\deg}}$, and the elevation uncertainty to be ${\sigma_\theta = \SI{0.1}{\deg}}$.  Note the azimuth angle is taken to lie along the laser line profile, whilst the elevation angle is perpendicular to the laser profile.  The covariance on the laser-to-INS extrinsic estimate is taken to be ${\mbs{\Sigma}^s_\ell = \diag\bigl( \sigma^2_\alpha \cdot \eye, \sigma^2_\beta \cdot \eye \bigr) \in \rnums^{6\times 6}}$, with ${\sigma_\alpha = \SI{0.1}{\deg}}$ and ${\sigma_\beta = \SI{5}{\milli\meter}}$.  Note that, in practice, this value would originate from an extrinsic calibration algorithm, with a higher variance assigned to calibration parameters with low observability \cite{Lv2022a,Hitchcox2024}.  

As raw IMU measurements are not available from the commercial DVL-INS, a white-noise-on-acceleration (WNOA) motion prior is used to model vessel kinematics \cite{Anderson2015,Hitchcox2023a}, 
\begin{equation}
    \mbfdot{T}^{zw}_{ab} = \mbf{T}^{zw}_{ab} \, \mbs{\varpi}_b^\wedge,
\end{equation}
in which ${\mbs{\varpi}_b = \left[ \mbs{\omega}_b^\trans \ \ \mbs{\nu}_b^\trans \right]^\trans \in \rnums^6}$ is the \textit{generalized velocity}, with ${\mbs{\omega}_b \in \rnums^3}$ the angular velocity and ${\mbs{\nu}_b \in \rnums^3}$ the linear velocity.  The continuous-time error kinematics are
\begin{equation}
	\underbrace{\begin{bmatrix}
		\delta \mbsdot{\xi} \\ \delta \mbsdot{\varpi}
	\end{bmatrix}}_{\delta \mbfdot{x}} =
	\underbrace{\begin{bmatrix}
		-\mbf{ad}(\mbsbar{\varpi}_b) & -\eye \\
		\mbf{0} & \mbf{0}
	\end{bmatrix}}_{\mbf{A}}
	\underbrace{\begin{bmatrix}
		\delta \mbs{\xi} \\ \delta \mbs{\varpi}
	\end{bmatrix}}_{\delta \mbf{x}} + 
	\underbrace{\begin{bmatrix}
		\mbf{0} \\ \eye
	\end{bmatrix}}_{\mbf{L}} \delta \mbf{w},
	\label{eqn:wnoacterrors}
\end{equation}
and together with ${\mbs{Q} = \diag \bigl( \sigma_{\dot{\omega}}^2 \cdot \eye, \sigma_{\dot{\nu}}^2 \cdot \eye \bigr) }$ the discrete-time process noise is given by \eqref{eqn:Qk}.  For illustrative purposes, the PSD on the WNOA parameters is set to ${\sigma^2_{\dot{\omega}} = \SI{9e-6}{\radian\squared\second\tothe{-3}}}$ and ${\sigma^2_{\dot{\nu}} = \SI{1e-8}{\meter\squared\second\tothe{-3}}}$, respectively.  \Cref{tab:noise_wiarton_noise_visualization_table} summarizes the parameters used for the Wiarton submap noise visualization.  

\begin{table}
    \centering
    \caption{Parameters for Wiarton covariance visualization}
    \renewcommand{\arraystretch}{1.2}
    \begin{tabularx}{\columnwidth}{X|M{1cm}M{2cm}}
    \toprule
    Parameter & Symbol & Value \\
    \hline
    Std. dev. range measurement & $\sigma_{\textrm{r}}$ & \SI{1e-2}{\meter} \\
    Std. dev. azimuth measurement & $\sigma_\phi$ & \SI{3}{\deg} \\
    Std. dev. elevation measurement & $\sigma_\theta$ & \SI{0.1}{\deg} \\
    Std. dev. extrinsic rotation & $\sigma_\alpha$ & \SI{0.1}{\deg} \\
    Std. dev. extrinsic translation & $\sigma_\beta$ & \SI{5e-3}{\meter} \\
    PSD WNOA rotational acc. & $\sigma^2_{\dot{\omega}}$ & \SI{9e-6}{\radian\squared\second\tothe{-3}} \\
    PSD WNOA translational acc. & $\sigma^2_{\dot{\nu}}$ & \SI{1e-8}{\meter\squared\second\tothe{-3}} \\
	\bottomrule
    \end{tabularx}
	\label{tab:noise_wiarton_noise_visualization_table}
\end{table}

\begin{figure}
	% the figure
	\centering
	\subcaptionbox{Noise envelopes grow larger away from the central pose (triad) \label{fig:noise_wiarton_top}}{%
	  \includegraphics[width=0.97\columnwidth]{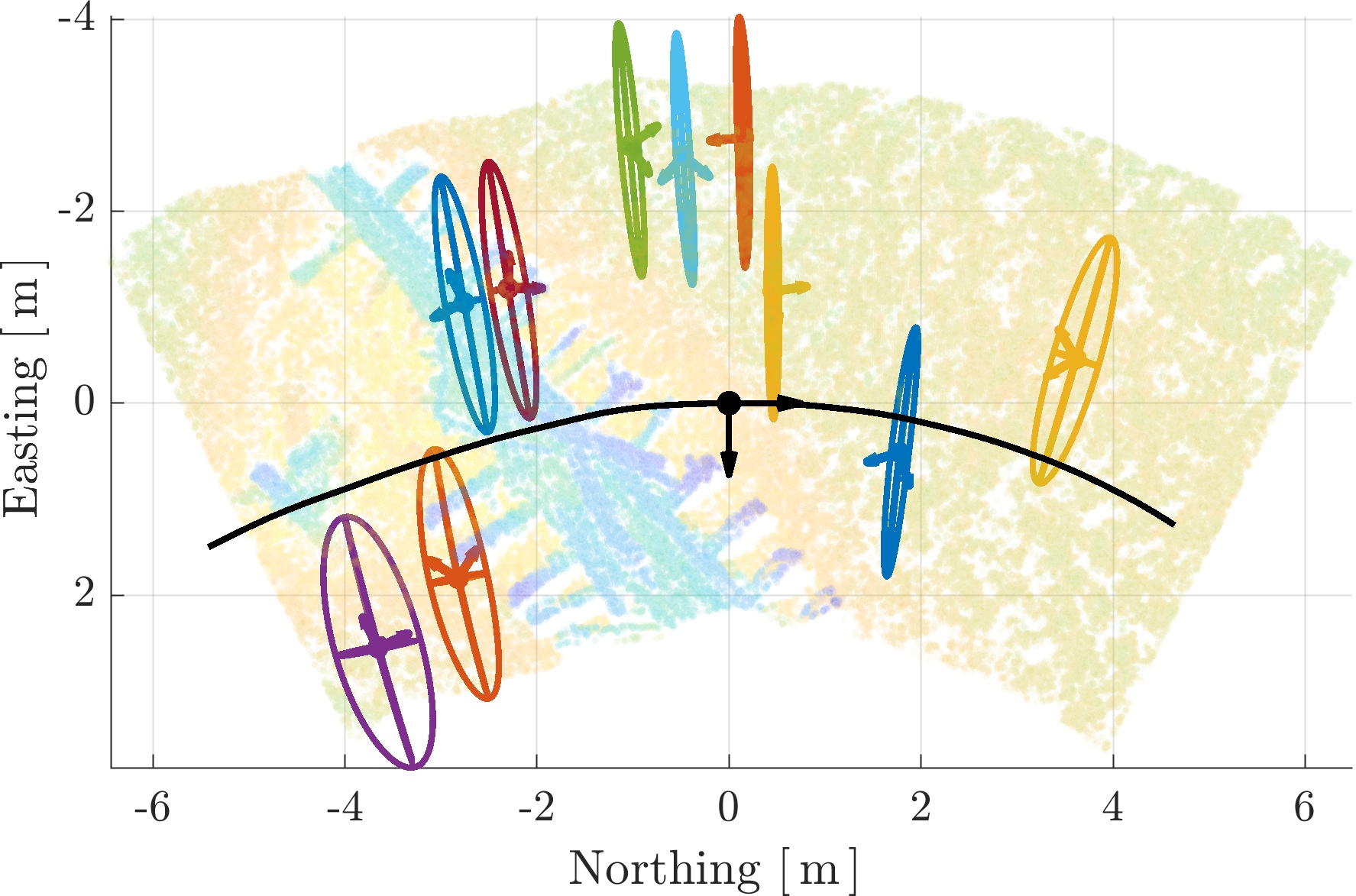}
	}
	\par \bigskip
	\subcaptionbox{An isometric view shows the ``banana'' shape of the envelopes \label{fig:noise_wiarton_iso}}{%
	  \includegraphics[width=0.97\columnwidth]{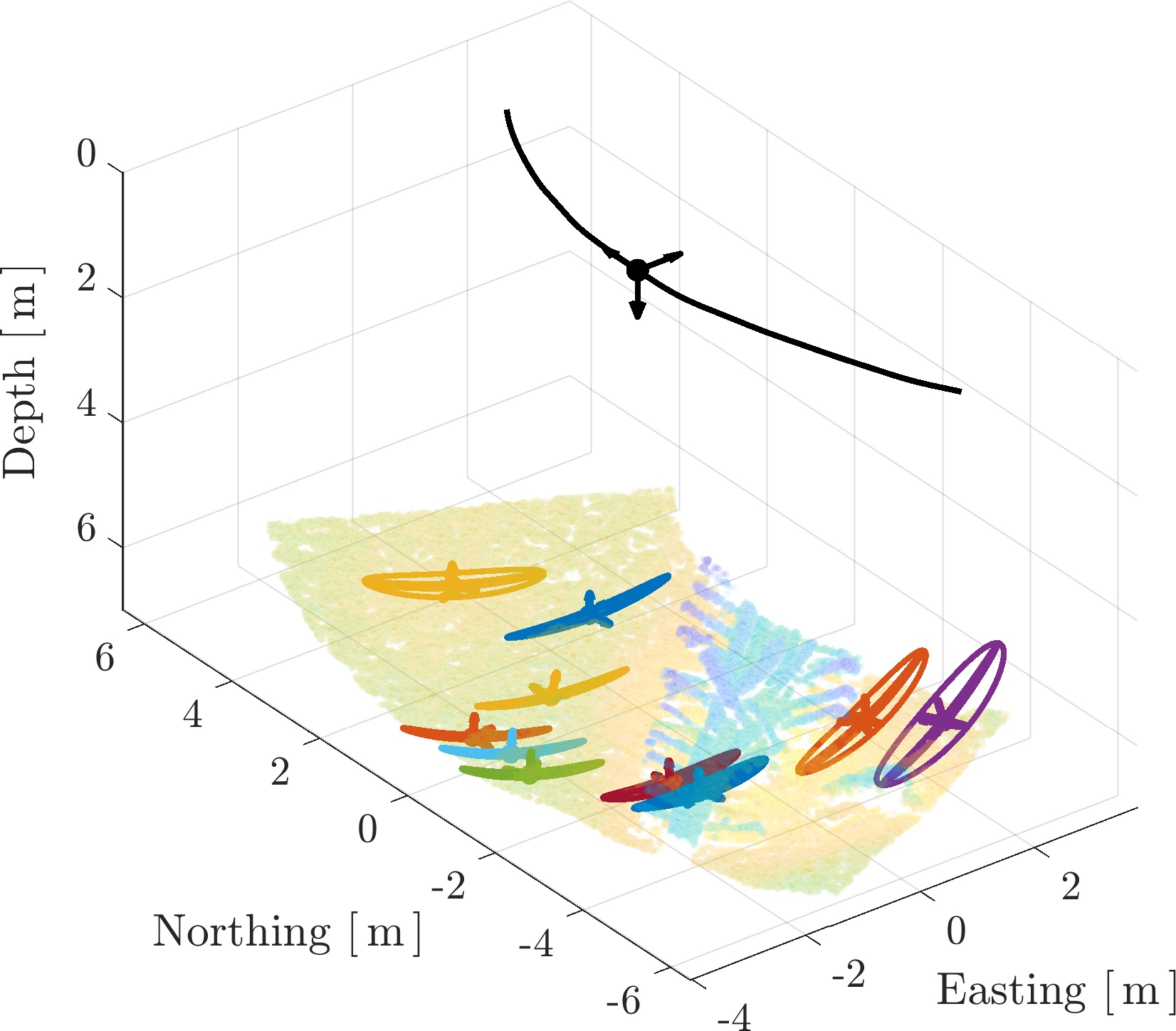}
	}
    \caption{\SI{99.73}{\percent} confidence envelopes for ten points randomly
    selected from a point-cloud submap containing a shipwreck.  The vehicle trajectory is shown in black, with the central submap pose shown as a black triad. }
    \label{fig:results_field}
\end{figure}

\Cref{fig:results_field} shows the resulting \SI{99.73}{\percent} confidence envelopes for ten points randomly selected from one submap.  The vehicle trajectory and central submap pose are shown in black.  Envelopes close to the central pose appear as slender bananas, reflecting the relatively large uncertainty in the azimuth direction along the laser lines.  As observed in the simulated example from \Cref{sec:results_sim}, the envelopes become larger and more ellipsoidal further away from the central pose, reflecting the influence of the WNOA odometry uncertainty on the submap noise profile.  Even a small amount of interoceptive uncertainty leads to a large variation in the noise profile across the submap domain.  This is sure to impact submaps collected during periods of dead-reckoning, especially when using lower-grade interoceptive sensors.  

% -----------------------------------------------------------
% -----------------------------------------------------------
\section{Conclusion}
\label{sec:conclusion}

Scan-matching is a widely-used technique in state estimation applications.  Modelling measurement uncertainty more accurately will lead to better scan-matching performance, both through better data association decisions and by providing for a more realistic uncertainty on the scan-matching constraint.  

Many sensors used for scan matching, for example radar, lidar, and laser scanners, may be fundamentally represented using a range-azimuth-elevation sensor model.  However, linearization of this model results in a poor approximation of the underlying noise characteristics.  Additional sources of uncertainty, such as sensor-to-vehicle extrinsic uncertainty and odometry uncertainty, are frequently overlooked or ignored when combining multiple measurements into point-cloud ``submaps'' for scan-matching applications.  

The main contributions of this paper are the development of an $SE(3)$ noise model for RAE sensors, and a straightforward methodology for incorporating extrinsic and odometry uncertainty into the submap noise profile.  Future work will focus on the real-world effects of interoceptive sensor bias, and on performance and consistency analysis in real-world scan matching applications.

% -----------------------------------------------------------
% -----------------------------------------------------------
\section*{Acknowledgment}

The authors would like to thank Ryan Wicks of Voyis for providing experimental
data and guidance.

% -----------------------------------------------------------
% -----------------------------------------------------------
% bibliography
\printbibliography

\end{document}